\pdfoutput=1

\documentclass[11pt]{article}

\usepackage{acl}

\usepackage{times}
\usepackage{latexsym}

\usepackage[T1]{fontenc}

\usepackage[utf8]{inputenc}

\usepackage{microtype}

\usepackage{inconsolata}
\usepackage{todonotes}
\usepackage{booktabs}
\usepackage{multirow}
\usepackage{enumitem}
\usepackage{makecell}

\usepackage{pifont}

\title{Event Detection from Social Media for Epidemic Prediction}

\author{Tanmay Parekh$^{\dagger}$ \ \ \
Anh Mac$^{\dagger}$ \ \ \
Jiarui Yu$^{\dagger}$ \ \ \
Yuxuan Dong$^{\dagger}$ \\
{\bf Syed Shahriar$^{\dagger}$ \ \ \
Bonnie Liu$^{\dagger}$ \ \ \
Eric Yang$^{\dagger}$ \ \ \
Kuan-Hao Huang$^{\mathsection}$} \\
{\bf Wei Wang$^{\dagger}$ \ \ \
Nanyun Peng$^{\dagger}$ \ \ \
Kai-Wei Chang$^{\dagger}$} \\
$^{\dagger}$Computer Science Department, University of California, Los Angeles \\
$^{\mathsection}$Department of Computer Science, University of Illinois Urbana-Champaign \\
\texttt{\{tparekh, weiwang, violetpeng, kwchang\}@cs.ucla.edu} \\
  }

\graphicspath{{images/}}

\begin{document}
\maketitle

\newcommand{\mypar}[1]{\vspace{0.35em}\noindent\textbf{#1}}
\newcommand{\SideNote}[2]{\todo[color=#1,size=\small]{#2}} 

\newcommand{\hb}[1]{\SideNote{blue!40}{#1 --hritik}}
\newcommand{\kuanhao}[1]{\SideNote{green!40}{#1 --kuan-hao}}
\newcommand{\tanmay}[1]{\SideNote{orange!40}{#1 --tanmay}}
\newcommand{\kaiwei}[1]{\SideNote{brown!40}{#1 --kai-wei}}
\newcommand{\violet}[1]{\SideNote{purple!40}{#1 --violet}}
\newcommand{\jiao}[1]{\SideNote{red!40}{#1 --Jiao}}
\newcommand{\wei}[1]{\SideNote{blue!40}{#1 --wei}}
\newcommand{\alex}[1]{\SideNote{yellow!40}{#1 --alex}}

\newcommand{\dataName}[0]{SPEED}
\newcommand{\mpoxDataName}[0]{DETECT-MPOX}

\newcommand{\cmark}{\ding{51}}%
\newcommand{\xmark}{\ding{55}}%
\newcommand{\tildemark}{\textbf{$\sim$}}%

\begin{abstract}

Social media is an easy-to-access platform providing timely updates about societal trends and events. Discussions regarding epidemic-related events such as infections, symptoms, and social interactions can be crucial for informing policymaking during epidemic outbreaks. In our work, we pioneer exploiting Event Detection (ED) for better preparedness and early warnings of any upcoming epidemic by developing a framework to extract and analyze epidemic-related events from social media posts. To this end, we curate an epidemic event ontology comprising seven disease-agnostic event types and construct a Twitter dataset \dataName{} with human-annotated events focused on the COVID-19 pandemic. Experimentation reveals how ED models trained on COVID-based \dataName{} can effectively detect epidemic events for three unseen epidemics of Monkeypox, Zika, and Dengue; while models trained on existing ED datasets fail miserably. Furthermore, we show that reporting sharp increases in the extracted events by our framework can provide warnings 4-9 weeks earlier than the WHO epidemic declaration for Monkeypox. This utility of our framework lays the foundations for better preparedness against emerging epidemics.


\end{abstract}

\section{Introduction}

Early warnings and effective control measures are among the most important tools for policymakers to be prepared against the threat of any epidemic \cite{DBLP:journals/bioinformatics/CollierDKGCTNDKTST08}.
World Health Organization (WHO) reports suggest that $65\%$ of the first reports about infectious diseases and outbreaks originate from informal sources and the internet \cite{heymann2001hot}.
Social media is an important information source here, as it is more timely than other alternatives like news and public health \cite{lamb-etal-2013-separating}, more publicly accessible than clinical notes \cite{DBLP:journals/jbi/LybargerOTY21}, and possesses a huge volume of content.\footnote{A daily average of 20 million tweets were posted about COVID-19 from May 15 -- May 31, 2020.}
This underscores the need for an automated system monitoring social media to provide early and effective epidemic prediction.

\begin{figure}[t]
    \centering
    \includegraphics[width=\columnwidth]{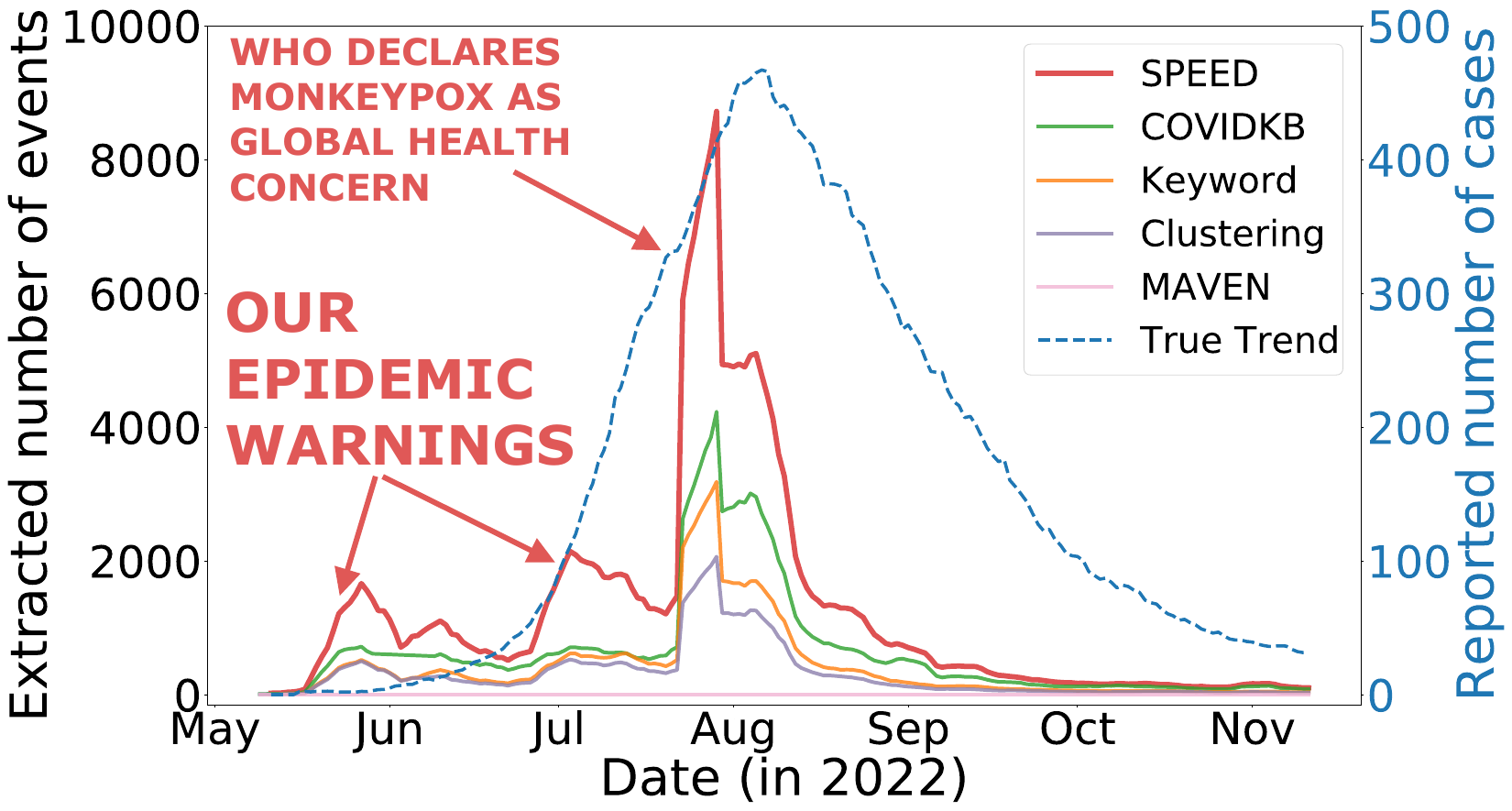}
    \caption{Number of reported Monkeypox cases and extracted events by our trained ED model from May 11 to Nov 11, 2022. Arrows indicate how our system could provide early epidemic warnings about 4-9 weeks before the WHO declared Monkeypox as a concern. We also show other baseline models for comparison.}
    \label{fig:mpox-total-trend}
\end{figure}

To this end, we pioneer to leverage the task of Event Detection (ED) for epidemic prediction.
ED involves identifying and categorizing significant events based on a pre-defined ontology \cite{sundheim-1992-overview, doddington-etal-2004-automatic}.
Compared to existing epidemiological keyword and sentence-classification approaches \cite{DBLP:journals/artmed/LejeuneBDL15, DBLP:journals/jbi/LybargerOTY21}, ED requires a deeper semantic understanding.
This enhanced understanding aids in more effective disease-agnostic extraction of epidemic events from social media.
By reporting sharp increases in extracted epidemic-related events, we can provide early epidemic warnings, as shown for Monkeypox in Figure~\ref{fig:mpox-total-trend} - highlighting the applicability of ED for epidemic prediction.

Existing ED datasets are unsuitable for establishing a framework to extract epidemic-related events from social media, as they focus on general-purpose events in news and wikipedia domains, while other epidemiological works are disease-specific and too fine-grained (\S~\ref{sec:related-works}).
Thus, we construct our own epidemic ED ontology and dataset for social media.
Our created ontology comprises seven event types - \textit{infect}, \textit{spread}, \textit{symptom}, \textit{prevent}, \textit{cure}, \textit{control}, \textit{death} - chosen based on their relevance for epidemics, frequency in social media, and their applicability to various diseases.
We further validate our ontology through clinical sources and public health experts.
For the dataset, we choose Twitter as the social media platform and focus on the COVID-19 pandemic.
Using our curated ontology and expert annotation, we create our dataset \textbf{\dataName{}} (\textbf{S}ocial \textbf{P}latform based \textbf{E}pidemic \textbf{E}vent \textbf{D}etection) comprising $1$,$975$ tweets and $2$,$217$ event mentions.
We complete our ED framework by training ED models \cite{du-cardie-2020-event, hsu-etal-2022-degree} on \dataName{}.
Overall, \dataName{} provides disease-agnostic coverage of epidemic events for social media; thus, serving as a valuable dataset for epidemic prediction.

To validate the utility of our ED framework for disease-agnostic epidemic prediction, we perform two evaluations for three unseen diseases Monkeypox, Zika, and Dengue.
First, we evaluate if our framework trained on our COVID-only \dataName{} dataset can detect epidemic events for the unseen diseases.
Experiments reveal that our framework can successfully extract epidemic events, providing gains up to $29\%$ F1 over the best few-shot model and $10\%$ F1 gain over supervised models trained on limited target disease data.

Our second evaluation validates if aggregation of our extracted events can provide early epidemic warnings.
Comparing our extracted events with the actual reported cases, we show that our framework can provide warnings up to 4-9 weeks earlier than the WHO declaration for the Monkeypox epidemic (Figure~\ref{fig:mpox-total-trend}).
Such early warnings aided with timely action can potentially lead to 2-4x reduction in the number of infections and deaths \cite{kamalrathne2023need}.
These results underscore the strong utility of our dataset and framework for upcoming epidemic prediction and preparedness.



The contribution of this work is threefold, first, we pioneer to utilize Event Detection to develop an effective framework capable of extracting events from social media and providing early warnings for any unforeseeable epidemic.
To support the proposed framework, our second contribution is the design of a disease-agnostic social-media tailored ontology and dataset \dataName{}.
Our final contribution is extensive experiments to demonstrate the inadequacy of existing methods and the substantial improvements achieved by models trained on \dataName{}.
This signifies the pivotal role of our dataset and framework in enhancing the efficacy of epidemic prediction.
We release the data, code, and trained models at \url{https://github.com/PlusLabNLP/SPEED}.



\section{From Event Detection to Epidemic Prediction}

Given a social media post, Event Detection (ED) \cite{sundheim-1992-overview, doddington-etal-2004-automatic} extracts and classifies significant events of interest.
By designing disease-agnostic epidemic-based events, we aim to train ED models to extract epidemic events from social media posts for any possible disease.
By detecting abnormal influx in the trends of extracted epidemic events from social media, we can thus provide early epidemic warnings for any possible disease, as we show for Monkeypox in Figure~\ref{fig:mpox-total-trend}.
Existing epidemiological approaches \cite{DBLP:journals/artmed/LejeuneBDL15, DBLP:journals/jbi/LybargerOTY21} are simple keyword or sentence classification-based and less accurate.
Other works like COVIDKB \cite{zong-etal-2022-extracting} and ExcavatorCovid \cite{min-etal-2021-excavatorcovid} are disease-specific and utilize events for building knowledge bases.
To the best of our knowledge, we are the first ones to leverage event detection to extract epidemic events from social media and provide early warnings for any possible disease.


\begin{figure}
    \centering
    \fbox{\includegraphics[width=\columnwidth]{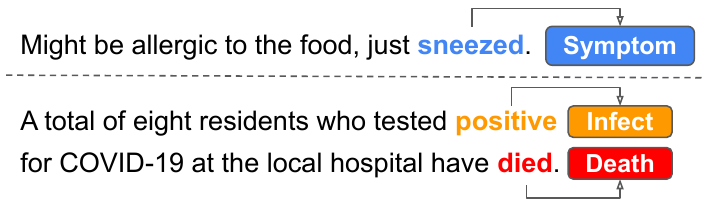}}
    \caption{Illustration for the task of Event Detection. Event mentions: Event \textit{symptom} and trigger \textit{sneezed} (1st sentence), Event \textit{infect} and trigger \textit{positive} (2nd sentence), Event \textit{death} and trigger \textit{died} (2nd sentence).}
    \label{fig:ed-example}
\end{figure}


\paragraph{Formal Task Definition}
Following ACE 2005 guidelines \cite{doddington-etal-2004-automatic},
we define an \textbf{event} to be something that happens or describes a change of state and is labeled by a specific \textbf{event type}.
An \textbf{event mention} is the sentence wherein the event is described.
Each event mention comprises an \textbf{event trigger}, which is the word/phrase that most distinctly highlights the occurrence of the event.
\textbf{Event Detection} is technically defined as the task of identifying event triggers from sentences and classifying them into one of the pre-defined event types (defined by an \textbf{event ontology}).
The subtask of identifying event triggers is called \textbf{Trigger Identification} and classification into event types is \textbf{Trigger Classification} \cite{ahn-2006-stages}.
Figure~\ref{fig:ed-example} shows examples for three event mentions for the events \textit{symptom}, \textit{infect}, and \textit{death}.

\section{Ontology Creation and Data Collection}


\noindent We choose social media as our document source as it provides faster and more timely worldly information than news and public health \cite{lamb-etal-2013-separating} and is more publicly accessible than clinical notes \cite{DBLP:journals/jbi/LybargerOTY21}.
Owing to its public access and huge content volume, we consider \textbf{Twitter}\footnote{\url{https://www.twitter.com/}} as the social media platform and consider the recent \textbf{COVID-19 pandemic} as the primary disease.

\begin{figure}[t]
    \centering
    \includegraphics[width=0.9\columnwidth]{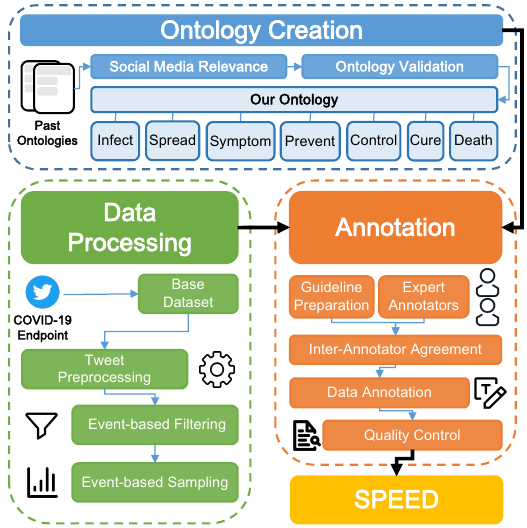}
    \caption{Overview of our dataset creation process with three major steps: Ontology Creation, Data Processing, and Data Annotation.}
    \label{fig:dataset-creation-process}
\end{figure}

Existing epidemiological ontologies are typically disease-specific, too fine-grained, or limited in coverage (\S~\ref{sec:epi-ontologies} and Table~\ref{tab:dataset-checklist}).
Similarly, standard ED datasets don't comprise epidemiological events and mostly focus on news or Wikipedia domains (\S~\ref{sec:ed-datasets}).
Due to these limitations, we create our own event ontology and dataset \dataName{} for detecting disease-agnostic epidemics from social media.
Figure~\ref{fig:dataset-creation-process} provides a brief overview of our data creation process, with further details discussed below.

\subsection{Ontology Creation}
\label{sec:ontology-creation}

Taking inspiration from medical sources like BCEO \cite{DBLP:journals/bioinformatics/CollierDKGCTNDKTST08}, IDO \cite{DBLP:journals/biomedsem/BabcockBCS21}, and the ExcavatorCovid \cite{DBLP:conf/emnlp/MinRQZXM21}, we curate a wide range of epidemic-related event types.
Next, we merge similar event types across these different ontologies (e.g. \textit{Outbreak} event type). 
To create a disease-agnostic ontology, we filter out event types biased for specific diseases (e.g. \textit{Mask Wearing} for COVID-19) and create disease-agnostic definitions using aid from public-health experts.
Finally, we categorize these events into three abstractions: personal (individual-oriented events), social (large population events), and medical (medically focused events) types.
We report our initial ontology comprising $18$ event types in Table~\ref{tab:complete-ontology} and share additional specifications in \S~\ref{sec:complete-ontology}.

\begin{table*}[ht]
    \centering
    \small
    \resizebox{.99\textwidth}{!}{
    \begin{tabular}{l|p{5.7cm}|p{8.4cm}}
    \toprule
    \textbf{Event Type} & \textbf{Event Definition} & \textbf{Example Event Mentions} \\
    \midrule
    \multirow{2}{*}{Infect} & \multirow{2}{*}{\shortstack[l]{The process of a disease/pathogen invading \\ host(s)}} & 1. Children can also \textbf{catch} COVID-19 ... \\ 
    & & 2. If you have antibodies, you \textbf{had} the virus. Period. \\ \midrule
    \multirow{2}{*}{Spread} & \multirow{2}{*}{\shortstack[l]{The process of a disease spreading or \\ prevailing massively at a large scale}} & 1. \#COVID-19 CASES \textbf{RISE} TO 85,940  IN INDIA ... \\ 
    & & 2. ... the \textbf{prevalence} of asymptomatic COVID - 19 cases ... \\ \midrule
    \multirow{3}{*}{Symptom} & \multirow{3}{*}{\shortstack[l]{Individuals displaying physiological features \\indicating the abnormality of organisms}} & 1. (user) (user) Still \textbf{coughing} two months after being infected by this stupid virus ... \\ 
    & & 2. If a person nearby is \textbf{sick}, the wind will scatter the virus ... \\ \midrule
    \multirow{2}{*}{Prevent} & \multirow{2}{*}{\shortstack[l]{Individuals trying to prevent the infection of \\ a disease}} & 1. ... wearing mask is the way to \textbf{prevent} COVID-19 \\ 
    & & 2. ... an \#antibody that has been succssful at \textbf{blocking} the virus \\ \midrule
    \multirow{2}{*}{Control} & \multirow{2}{*}{\shortstack[l]{Collective efforts trying to impede the spread \\ of an epidemic}} & 1. Social Distancing \textbf{reduces} the spread of covid ... \\ 
    & & 2. (user) COVID is still among us! Wearing masks \textbf{saves} lives! \\ \midrule
    \multirow{2}{*}{Cure} & \multirow{2}{*}{\shortstack[l]{Stopping infection and relieving individuals \\ from infections/symptoms}} & 1. ... \textbf{recovered} corona virus patients cant get it again \\
    & & 2. ... patients are \textbf{treated} separately at most places \\ \midrule
    \multirow{2}{*}{Death} & \multirow{2}{*}{\shortstack[l]{End of life of individuals due to an \\ infectious disease}} & 1. More than 80,000 Americans have \textbf{died} of COVID ... \\
    & & 2. The virus is going to get people \textbf{killed}. Stay home. Stay safe. \\
    \bottomrule
    \end{tabular}}
    \caption{Event ontology comprising seven event types promoting epidemic preparedness along with their definitions and two example event mentions. The trigger words are marked in \textbf{bold}.}
    \label{tab:event-ontology}
\end{table*}

\mypar{Social Media Relevance}
To tailor our curated ontology for social media, we conduct a deeper analysis of the event types based on their frequency and specificity.
Our goal is to filter and merge event types that occur less frequently and less distintively in social media.
To this end, using human expertise and external tools like Thesaurus,\footnote{\url{https://www.thesaurus.com/}} we first associate each event type with specific keywords.
Then we rank the event types based on the specificity and frequency of their keywords in social media posts.
Based on this ranking, we merge and discard the lower ranked event types (e.g. \textit{Respond} and \textit{Prefigure}).
Furthermore, we conduct human studies and merge event types to ensure better pairwise distinction (e.g. \textit{Treatment} is merged with \textit{Cure}).
We provide the keywords associated with each event type and the example social media posts in Table~\ref{tab:event-keywords-posts} along with additional details in \S~\ref{sec:initial-event-analysis}.

\mypar{Ontology Validation and Coverage}
Elemental medical soundness is ensured for our ontology since it is derived from established epidemiological ontologies.
To further certify this soundness, two public health experts (epidemiologists working in the Department of Public Health) validate the sufficiency and comprehensiveness of our ontology and event definitions.
To verify if our ontology is characteristic of any disease, we assess our ontology coverage for four diverse diseases by estimating the percentage of event occurrence in disease-related tweets.
Notably, we observe a coverage of $50\%$ for COVID-19, $44\%$ for Monkeypox, $70\%$ for Dengue, and $73\%$ for Zika (details in \S~\ref{sec:disease-coverage}).
Our coverage is better than the coverages of standard ED datasets ACE \cite{doddington-etal-2004-automatic} (27\%) and ERE \cite{song-etal-2015-light} (43\%), in turn, confirming the robust disease coverage of our ontology.

Our final ontology comprises seven primary event types tailored for social media, disease-agnostic, and encompassing crucial aspects of an epidemic.
We present our ontology in Table~\ref{tab:event-ontology} along with event definitions and example event mentions.

\subsection{Data Processing}
\label{sec:twitter-selection}

To access a wide range of tweets related to COVID-19, we utilized the Twitter COVID-19 Endpoint released in April 2020.
We used a randomized selection of \textbf{331 million tweets} between May 15 -- May 31 2020, as our base dataset.
For preprocessing tweets, we follow \citet{DBLP:journals/eswa/PotaVFE21}: (1) we anonymize personal information like phone numbers, emails, and handles, (2) we normalize any retweets and URLs, (3) we remove emojis and split hashtags, (4) we filter out tweets only in English.

\mypar{Event-based Filtering}
Most tweets in our base dataset expressed subjective sentiments, while only 3\% comprised mentions aligned with our event ontology.\footnote{Based on keyword-based study conducted on 1,000 tweets}
To reduce annotation costs, we further filter these tweets 
using a simple \textit{sentence embedding} similarity technique.
Specifically, each event type is linked to a seed repository of $5$-$10$ diverse tweets.
Example seed tweets per event are shown in Table~\ref{tab:seed-tweets} in \S~\ref{sec:seed-tweets}.
Query tweets are filtered based on their sentence-level similarity \cite{reimers-2019-sentence-bert} with this event-based seed repository.\footnote{We use a filtering threshold of $0.9$.} 
This step filters about $95\%$ tweets from our base dataset, leading to 20x reduction in the annotation cost. 

\mypar{Event-based Sampling}
\label{sec:event-based-sampling}
Random sampling of tweets would yield an uneven and COVID-biased distribution of event types for our dataset.
We instead perform a uniform sampling - wherein we over-sample tweets linked to less frequent types (e.g. \textit{prevent}) and under-sample the more frequent ones (e.g. \textit{death}).
Such a uniform sampling has proven to ensure model robustness \cite{parekh-etal-2023-geneva} - as also validated by our experiments (\S~\ref{sec:uniform-sampling}) - and in turn, would make \dataName{} generalizable to a wider range of diseases.
In total, we sample $1$,$975$ tweets which are utilized for ED annotation.

\subsection{Data Annotation}
\label{sec:data-annotation}

For ED annotation, annotators are tasked with identifying whether a given tweet mentions any event outlined in our ontology.
If an event is present, annotators are required to identify the specific event trigger.
We design our annotation guidelines following the standard ACE dataset \cite{doddington-etal-2004-automatic} and amend them through several rounds of preliminary annotations to ensure annotator consistency.
Additional details are provided in \S~\ref{sec:annotation-guidelines-interface}.

\mypar{Annotator Details}
To ensure high annotation quality and consistency, we chose six experts instead of crowdsourced workers.
These experts are computer science students studying NLP and well-versed for ED.
They were further trained through multiple rounds of annotations and feedback.

\mypar{Inter-annotator agreement (IAA)}
We used Fleiss' Kappa \cite{fleiss1971measuring} for measuring IAA.
We conduct two phases of IAA studies:
(1) \textit{Guideline Improvement:} Three annotators participated in three annotation rounds to improve the guidelines through collaborative discussion of disagreements. IAA score rose from $0.44$ in the first round to $0.59$ ($70$ samples) in the final round.
(2) \textit{Agreement Improvement:} All annotators participated in three rounds of annotations to boost consistency. IAA score improved from $0.56$ in the first round to a strong $0.65$ ($50$ samples) in the final round.

\mypar{Quality Control}
We further ensure high annotation quality through: 
(1) \textit{Multi-Annotation:} each tweet is annotated by two annotators, disagreements resolved by a third, and
(2) \textit{Flagging:} annotators flag ambiguous annotations, resolved by a third annotator via discussion. These, coupled with good IAA scores, ensure the high quality of our annotations.

\subsection{Data Analysis}
\label{sec:data-analysis}


Our dataset \dataName{} comprises seven event types with $2$,$217$ event mentions annotated over $1$,$975$ tweets.
We compare \dataName{} with other ED datasets like ACE \cite{doddington-etal-2004-automatic}, ERE \cite{song-etal-2015-light}, M$^2$E$^2$ \cite{li-etal-2020-cross}, MLEE \cite{DBLP:journals/bioinformatics/PyysaloOMCTA12}, FewEvent \cite{DBLP:conf/wsdm/DengZKZZC20}, and MAVEN \cite{wang-etal-2020-maven} in Table~\ref{tab:data-statistics}.
We show how other datasets focus on the news, biomedical, general, and Wikipedia domains, while \dataName{} is the first-ever ED dataset for social media, specifically Twitter.
Furthermore, none of the previous datasets comprise any of the epidemiological event types present in \dataName{} (\S~\ref{sec:event-coverage}).


\begin{table}[t]
    \centering
    \small
    \setlength{\tabcolsep}{.5pt}
    \begin{tabular}{lccccc}
        \toprule
        \multirow{2}{*}{\textbf{Dataset}} & \textbf{\# Event} & \multirow{2}{*}{\textbf{\# Sent}} & \multirow{2}{*}{\textbf{\# EM}} & \textbf{Avg. EM} & \multirow{2}{*}{\textbf{Domain}} \\
        & \textbf{Types} & & & \textbf{per Event} \\
        \midrule
        \textbf{ACE} & $33$ & $18,927$ & $5,055$ & $153$ & News \\
        \textbf{ERE} & $38$ & $17,108$ & $7,284$ & $192$ & News \\
        \textbf{M}$^2$\textbf{E}$^2$ & $8$ & $6,013$ & $1,105$ & $138$ & News \\
        \textbf{MLEE} & $29$ & $286$ & $6,575$ & $227$ & Biomedical \\
        \textbf{FewEvent} & $100$ & $12,573$ & $12,573$ & $126$ & General \\
        \textbf{MAVEN} & $168$ & $49,873$ & $118,732$ & $\textbf{707}$ & Wikipedia \\
        \textbf{\dataName} & $7$ & $1,975$ & $2,217$ & $\textbf{317}$ & Social Media \\
        \bottomrule
    \end{tabular}
    \caption{Data Statistics for \dataName{} dataset and comparison with other standard ED datasets. \# = ``number of", Avg. = average, Sent = sentences, EM = event mentions.}
    \label{tab:data-statistics}
\end{table}

\begin{figure}[t]
    \centering
    \includegraphics[width=\columnwidth]{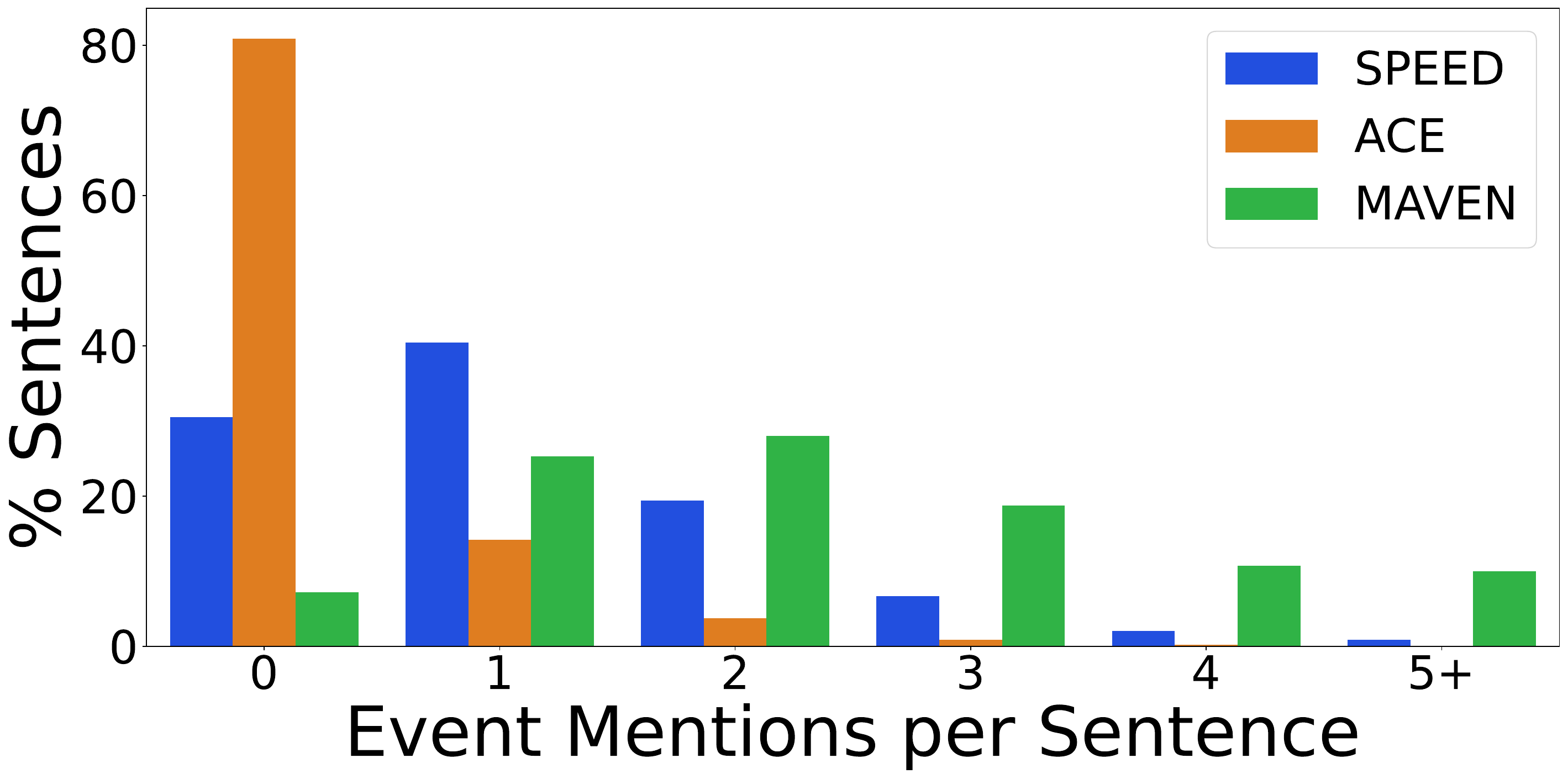}
    \caption{Distribution of number event mentions per sentence. Here \% indicates percentage.}
    \label{fig:event-mention-distribution}
\end{figure}

\paragraph{Comparable Datasize}
Since we only focus on $7$ event types, \dataName{} has relatively lesser number of sentences and event mentions.
However, \dataName{} has a high $316$ average mentions per event type (column 5 in Table~\ref{tab:data-statistics}), more than most other standard datasets.
We compare the distribution of event mentions per sentence with other ED datasets like ACE and MAVEN in Figure~\ref{fig:event-mention-distribution}.
We observe that the event density of our dataset is less than MAVEN but better than ACE.
This shows that \dataName{} is fairly dense and reasonably sized ED dataset.

\paragraph{Diverse and Challenging}
We show the diversity of trigger words in \dataName{} and compare it with other datasets in Table~\ref{tab:trigger-word-diversity}.
We note that \dataName{} has a strong average number of triggers per event mention.
This demonstrates how \dataName{} is a diverse and challenging ED dataset.

\begin{table}[t]
    \centering
    \small
    \begin{tabular}{lrrr}
        \toprule
        \multirow{2}{*}{\textbf{Dataset}} &  \textbf{\#Event} & \textbf{\# Unique} & \textbf{Avg. Triggers} \\
        & \textbf{Mentions} & \textbf{Triggers} & \textbf{per Mention} \\
        \midrule
        ACE & $5,055$ & $1,229$ & $0.24$ \\
        MAVEN & $118,732$ & $7,074$ & $0.06$ \\
        \dataName{} & $1,975$ & $555$ & $\textbf{0.25}$ \\
        \bottomrule
    \end{tabular}
    \caption{Comparison of \dataName{} with ACE and MAVEN in terms of unique trigger words and average number of triggers per event mention. Avg = Average.}
    \label{tab:trigger-word-diversity}
\end{table}










\section{Epidemic Prediction}
\label{sec:new-pandemic}




For our ED framework, we utilize our curated dataset \dataName{} to train various ED models (\S~\ref{sec:ed-models}).
To validate the utility of models for the application of epidemic prediction, we perform evaluations using two tasks: (1) Epidemic event detection and (2) Early warning prediction.
Epidemic event detection performs a formal ED evaluation of the models for detecting epidemic-based events.
On the other hand, early warning prediction practically evaluates if the extracted events by the model can be aggregated to provide any early epidemic warnings.

Since \dataName{} focuses solely on COVID-19, we conduct these epidemic prediction evaluations for three unseen epidemics of \textit{Monkeypox} (2022), \textit{Zika} (2017), and \textit{Dengue} (2018).
These diseases are fairly distinct too, as Monkeypox causes rashes and rarely fatal, Zika causes birth defects, and Dengue causes high fever and can be fatal.
For our evaluations, we utilize and modify the raw Twitter dumps provided by \citet{DBLP:data/10/Thakur22b} for Monkeypox and \citet{Dias2020} for Zika and Dengue.



\subsection{Epidemic Event Detection}

To validate if our \dataName{}-trained models can extract events for any epidemic, we perform traditional ED evaluation of these models for unseen diseases of Monkeypox, Zika, and Dengue.
Following \citet{ahn-2006-stages}, we report the F1-score for trigger identification (\textbf{Tri-I}) and classification (\textbf{Tri-C}).

\begin{table}[t]
    \centering
    \small
    \begin{tabular}{l|rrr}
        \toprule
        & \textbf{Disease} & \textbf{\# Sent} & \textbf{\# EM} \\
        \midrule
        \textbf{Train} & COVID & $1,601$ & $1,746$ \\ \midrule
        \textbf{Dev} & COVID & $374$ & $471$ \\ \midrule
        \multirow{2}{*}{\textbf{Test}} & Monkeypox & $286$ & $398$ \\
        & Zika + Dengue & $300$ & $274$ \\
        \bottomrule
    \end{tabular}
    \caption{Statistics for data splits for epidemic event detection evaluation. \# = ``number of", Sent = sentences, EM = event mentions.}
    \label{tab:data-setup}
\end{table}

\paragraph{Data Setup}
To train our ED models, we split the \dataName{} into 80-20 split for training and development sets.
For testing, we sample tweets from the Twitter dumps of Monkeypox, Zika, and Dengue.
Since the original data doesn't comprise event-based annotations, we utilize the same human experts who annotated SPEED to annotate the raw tweets for ED and create the evaluation dataset.
We provide statistics for our data setup in Table~\ref{tab:data-setup}.
We release the training COVID data along with this evaluation data for Zika, Monkeypox, and Dengue for future benchmarking.

\paragraph{ED Models}
\label{sec:ed-models}
For training models using \dataName{} for our ED framework, we consider the following supervised models:
(1) DyGIE++ \cite{wadden-etal-2019-entity},
(2) BERT-QA \cite{du-cardie-2020-event},
(3) DEGREE \cite{hsu-etal-2022-degree},
(4) TagPrime \cite{hsu-etal-2023-simple}.
We utilized the TextEE framework \cite{DBLP:journals/corr/abs-2311-09562} to implement these models and provide more details in \S~\ref{sec:implementation-details}.

\paragraph{Baseline Models}
As baselines, we consider zero-shot ED models (\textbf{\textsc{Zero-shot}}) that do not train on any supervised data and solely utilize the event definitions.
We consider the following zero-shot models:
(1) TE \cite{lyu-etal-2021-zero},
(2) WSD \cite{yao-etal-2021-connect},
(3) ETypeClus \cite{shen-etal-2021-corpus}. Additional model implementation details is provided in \S~\ref{sec:implementation-details}.
We also consider transferring from existing datasets (\textbf{\textsc{Transfer from Existing Datasets}}) by training models on standard ED datasets like ACE \cite{doddington-etal-2004-automatic} and MAVEN \cite{wang-etal-2020-maven} without fine-tuning on epidemic ED data.

As stronger baselines, we also consider models utilizing epidemic ED data.
Here, we consider models using few-shot target disease data without any model training (\textbf{\textsc{No Training}}) like:
(1) Keyword \cite{DBLP:journals/artmed/LejeuneBDL15}, an epidemiological model utilizing curated event-specific keywords to detect events, and
(2) GPT-3.5 \cite{DBLP:journals/corr/abs-2005-14165}, a large-language model (LLM) using GPT-3.5-turbo with seven target disease in-context ED examples.
Finally, we consider super-strong baselines training ED models on limited target epidemic data (\textbf{\textsc{Trained on Target Epidemic}}).
Specifically, we sample 300 event mentions for the target disease (exclusive from the test data) and use human experts to annotate these tweets (as done for COVID annotations) to create the target epidemic data.
Note that these models are added for comparison, but they are practically infeasible for epidemic prediction, as it takes 3-6 weeks after the first infection to collect such target disease data.

\begin{table}[t]
    \centering
    \small
    \setlength{\tabcolsep}{4pt}
    \begin{tabular}{lcc|cc}
        \toprule
        \multirow{2}{*}{\textbf{Model}} & \multicolumn{2}{c|}{\textbf{Monkeypox}} & \multicolumn{2}{c}{\textbf{Zika + Dengue}} \\
        & \textbf{Tri-I} & \textbf{Tri-C} & \textbf{Tri-I} & \textbf{Tri-C}\\
        \midrule
        \multicolumn{5}{c}{\small \textsc{Zero-shot}} \\
        \midrule
        TE & $16.70$ & $12.11$ & $12.69$ & $9.06$ \\
        WSD & $22.04$ & $4.35$ & $27.93$ & $5.85$ \\
        ETypeClus & $18.31$ & $6.78$ & $13.99$ & $5.33$ \\
        \midrule
        \multicolumn{5}{c}{\small \textsc{Transfer from Existing Datasets}} \\ 
        \midrule
        ACE - TagPrime & $4.80$ & $0$ & $23.64$ & $0$ \\
        ACE - DEGREE & $12.15$ & $5.14$ & $14.47$ & $0$ \\
        MAVEN - TagPrime & $29.16$ & $0$ & $33.97$ & $0$ \\
        MAVEN - DEGREE & $27.94$ & $0$ & $32.04$ & $0$ \\
        \midrule
        \multicolumn{5}{c}{\small \textsc{No Training}} \\
        \midrule
        Keyword & $36.40$ & $25.09$ & $25.93$ & $21.69$ \\
        GPT-3.5 & $42.23$ & $35.33$ & $53.22$ & $14.27$ \\
        \midrule
        \multicolumn{5}{c}{\small \textsc{Trained on Target Epidemic}} \\
        \midrule
        BERT-QA & $59.8$ & $54.08$ & $64.99$ & $57.02$ \\
        DEGREE & $59.58$ & $54.12$ & $67.34$ & $62.9$ \\
        TagPrime & $55.57$ & $49.65$ & $70.92$ & $63.93$ \\
        DyGIE++ & $55.83$ & $50.31$ & $\textbf{73.24}$ & $\textbf{65.65}$ \\
        \midrule
        \multicolumn{5}{c}{\small \textsc{Trained on \dataName{} (Our Framework)}} \\
        \midrule
        BERT-QA & $\textbf{67.38}$ & $\textbf{64.17}$ & $67.95$ & $56.81$ \\
        DEGREE & $62.95$ & $61.45$ & $64.14$ & $56.63$ \\
        TagPrime & $64.71$ & $61.92$ & $62.08$ & $54.47$ \\
        DyGIE++ & $62.76$ & $59.82$ & $66.4$ & $56.4$ \\
        \bottomrule
    \end{tabular}
    \caption{Evaluating ED models trained on \dataName{} for detecting events for new epidemics of Monkeypox, Zika, and Dengue in terms of F1 scores.}
    \label{tab:generalizability-results}
\end{table}

\paragraph{Results}
We present our results in Table~\ref{tab:generalizability-results}.
Firstly, none of the existing data transfer, zero-shot, or no training-based models perform well for our task, mainly owing to the domain shift of social media and unseen epidemic events.
Overall, ED models trained on \dataName{} perform the best,
thus \textbf{demonstrating the capability of our ED framework to detect epidemic events for new diseases}.
Compared with models trained on the target epidemic, \dataName{}-trained models provide a gain of $10$ F1 points for Monkeypox and a reasonable performance for Zika and Dengue. 
This outcome is particularly encouraging, as it \textbf{demonstrates the resilience of our framework, making it highly applicable during the early stages of an epidemic, when minimal to no epidemic-specific data is accessible}.


\subsection{Early Warning Prediction}


As a practical validation of the utility of our framework, we evaluate if \dataName{}-trained ED models are capable of providing early warnings for an unknown epidemic.
More specifically, we aggregate the extracted event mentions by our framework over a time period and report any sharp increase in the rolling average of detected events as an epidemic warning.
For evaluation, we compare it with the actual number of disease infections reported in the same period.
Naturally, the earlier we provide an epidemic warning, the better the framework is deemed.
For this evaluation, we choose Monkeypox as the unseen disease and its outbreak from May 11 to Nov 11, 2022, as the unknown period.
We utilize the Twitter dump by \citet{DBLP:data/10/Thakur22b} as the datasource.
We also include four other baselines for comparison:
(1) MAVEN, an ED model trained on existing ED dataset MAVEN \cite{wang-etal-2020-maven},
(2) Clustering, a zero-shot classification baseline based on ETypeClus \cite{shen-etal-2021-corpus},
(3) Keyword, a no-training baseline based on a previous epidemiological model \cite{DBLP:journals/artmed/LejeuneBDL15},
(4) COVIDKB, a BERT-based classification model trained on a previous COVIDKB \cite{zong-etal-2022-extracting} dataset.

\paragraph{Results}
We report the number of epidemic events extracted by the BERT-QA trained on \dataName{} along with the actual number of Monkeypox cases reported in the US\footnote{As reported by CDC at \url{https://www.cdc.gov/poxvirus/mpox/response/2022/mpx-trends.html}} from May 11 to Nov 11, 2022, in Figure~\ref{fig:mpox-total-trend}.
As indicated by the arrows, our model could potentially provide two sets of early warnings around May 23 (9 weeks earlier, when first cases were detected) and June 29 (4 weeks earlier, when cases started rising) before the outbreak reached its peak around July 30.
Comparatively, MAVEN-trained model fails completely, while the clustering, keyword, and COVIDKB models have super weak trends to provide any kind of warnings.
In fact, all ED models trained on our SPEED data are capable of providing these early signals as shown in Figure~\ref{fig:mpox-4-ED-models} (further event-wise analysis in Appendix~\ref{sec:mpox-trends-appendix}).
To further validate if the early warnings are raised owing to some actual Monkeypox epidemic discussion, we show some tweets classified as epidemic events by our system in Table~\ref{tab:mpox-actual-preds}.
These tweets demonstrate that our system is picking the right signal to provide early epidemic warnings.
Overall, this robust outcome underscores the \textbf{practical utility of our framework to provide early epidemic warnings and ensure better preparedness for any potential epidemic.}


\begin{figure}[t]
    \centering
    \includegraphics[width=\columnwidth]{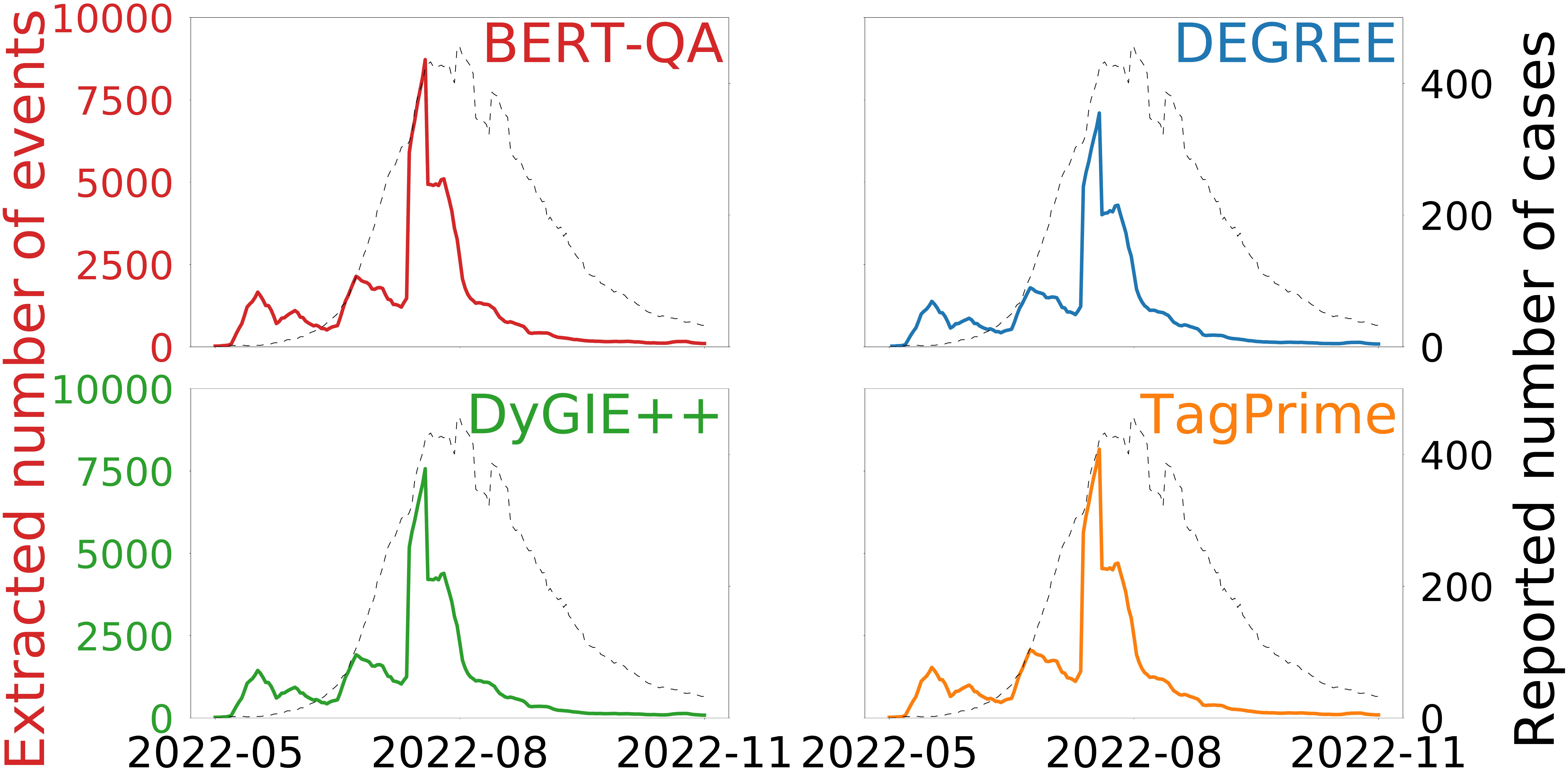}
    \caption{Number of reported Monkeypox cases and the number of extracted events from our four trained models from May 11 to Nov 11, 2022.}
    \label{fig:mpox-4-ED-models}
\end{figure}

\begin{table}[t]
    \centering
    \small
    \begin{tabular}{p{7.2cm}}
        \toprule
        The first known \textbf{case} {\color{blue}[\textit{infect}]} of human to dog \#(monkey pox) \textbf{transmission} {\color{blue}[\textit{spread}]} has been documented in medical journal, the Lancet. The dog \textbf{caught} {\color{blue}[\textit{infect}]} it from its owners, a non-monogamous gay couple in Paris. The greyhound \textbf{developed} {\color{blue}[\textit{symptom}]} an anal ulceration \& mucocutaneous lesions. (url) \\ \hline
        (user) Release Topxx for \textbf{treatment} {\color{blue}[\textit{cure}]} of monkey pox people are suffering \\ \hline
        (user) Wait what? I thought we were going to get Monkey Pox \textbf{vaccine} {\color{blue}[prevent]} in October. \\ \hline
        Monkey Pox is not an STI. It is not an STI. \#(monkey pox is not an sti)  It can be \textbf{transferred} {\color{blue}[spread]} via sex the same way Covid can. \\ \hline
        (user) Most of those people \textbf{have} {\color{blue}[infect]} Monkey Pox now. \\ \hline
        (user) More \textbf{dead} {\color{blue}[death]} in Chicago over a weekend than the world wide health emergency monkey pox \\
        \bottomrule
    \end{tabular}
    \caption{Sample illustrations of various tweets (anonymized) predicted as Monkeypox epidemic-related by our ED framework. The triggers are highlighted in \textbf{bold} with the respective event names in {\color{blue}[]}.}
    \label{tab:mpox-actual-preds}
\end{table}

\section{Analysis and Discussion}

In this section, we provide additional analyses to support the utility of our ED framework and also provide intuitions for why our framework works better than previous works.

\begin{table*}[t]
    \small
    \centering
    \begin{tabular}{lrp{5cm}p{5.8cm}}
    \toprule
         & \textbf{Disease} & \textbf{Infect Event Example} & \textbf{Symptom Event Example} \\
    \midrule
        Keyword-based & COVID-19 & Three students infected with \textbf{COVID-19} & COVID-19 symptoms include \textbf{fever}, \textbf{cough}, ... \\
        Keyword-based & Monkeypox & How do you catch \textbf{Monkeypox}? & Monkeypox may cause \textbf{rashes} and \textbf{itching} ... \\
        \dataName{} (Ours) & COVID-19 & Three students \textbf{infected} with COVID-19 & COVID-19 \textbf{symptoms} include fever, cough, ... \\
        \dataName{} (Ours) & Monkeypox & How do you \textbf{catch} Monkeypox? & Monkeypox may \textbf{cause} rashes and itching ... \\
    \bottomrule
    \end{tabular}
    \caption{Qualitative analysis for annotation difference between previous keyword-based epidemiological datasets \cite{DBLP:journals/bioinformatics/CollierDKGCTNDKTST08, DBLP:journals/artmed/LejeuneBDL15} and \dataName{}'s Event Detection based annotation schema. Our annotation schema is less disease-specific and thus, better generalizable to a wide range of diseases.}
    \label{tab:why-speed-generalize}
\end{table*}

\subsection{Event-based Disease Profiling}
\label{sec:disease-profiling}

\begin{figure}[t]
    \centering
    \includegraphics[width=\columnwidth]{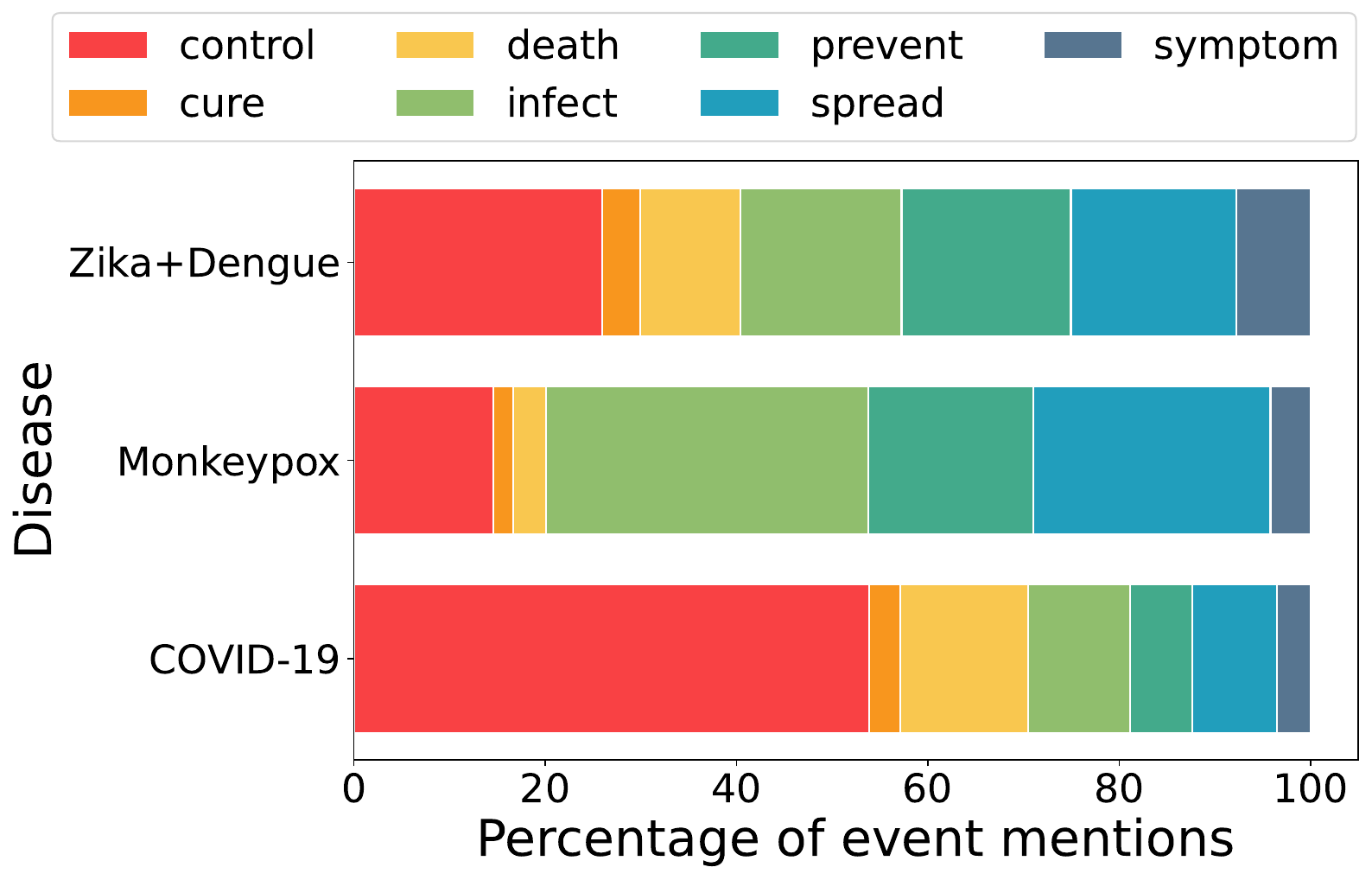}
    \caption{Disease profiles of public opinions generated by plotting the percentage of extracted event mentions for COVID-19, Monkeypox and Zika.}
    \label{fig:disease-profiling}
\end{figure}

Our ED framework offers the additional utility of generating event-based disease profiles using public sentiments.
These disease profiles can be generated by plotting the percentage of mentions per event type extracted by our framework.
Using 500k tweets, we depict the profiles for COVID, Monkeypox, and Zika+Dengue in Figure~\ref{fig:disease-profiling}.

Distinctive profiles emerge for each disease; COVID majorly comprises \textit{control}, Monkeypox exhibits a bias toward \textit{infect} and \textit{spread}, while Zika+Dengue emphasizes \textit{control} and \textit{death}.
These trends align with the higher fatality rate of Zika and Dengue \cite{paixao2022mortality}, recent discoveries of transmission routes of Monkeypox \cite{kozlov2022deadly}, and the need for mass public control measures for the COVID pandemic \cite{guner2020covid}.
Relatively, Monkeypox also shows low mentions for \textit{death}, \textit{cure} - which aligns with low fatality and no available cure for Monkeypox \cite{kmiec2022monkeypox}. 
Overall, these profiles can provide policymakers with valuable insights about new unknown outbreaks to implement more informed and effective interventions.


\subsection{Why does \dataName{} generalize?}
\label{sec:discussion-generalize}


We provide a qualitative analysis of why COVID-based \dataName{} helps detect epidemic events for other unseen diseases compared to previous epidemiological works \cite{DBLP:journals/bioinformatics/CollierDKGCTNDKTST08, DBLP:journals/artmed/LejeuneBDL15} and attribute it to the difference in the task formulation and annotation schema.
We demonstrate this difference (highlighted in \textbf{bold}) through illustrative examples for \textit{Infect} and \textit{Symptom} events in Table~\ref{tab:why-speed-generalize}.
As evident, keyword-based modeling requires annotating highly precise but disease-specific keywords like \textit{COVID-19}, \textit{fever}, etc.
On the other hand, our ED annotation formulation emphasizes the annotation of disease-agnostic triggers like \textit{infected}, \textit{symptoms}, etc.
This provides \dataName{} and our framework superior generalizability without new annotation to unseen diseases.

\section{Related Work}
\label{sec:related-works}

\begin{table}[t]
    \centering
    \small
   \setlength{\tabcolsep}{2pt}
    \begin{tabular}{lrrrrrrr}
        \toprule
        \multirow{2}{*}{\textbf{Dataset}} &  & \textbf{Sent-} &  & \textbf{Social} & \textbf{Per.} &  \\
        & \textbf{Source} & \textbf{Level} & \textbf{Trig.} & \textbf{Eve.} & \textbf{Eve.} & \textbf{SMG} \\
        \midrule
        \dataName{} (Ours) & Twitter & \cmark & \cmark & \cmark & \cmark & \cmark \\
        COVIDKB & Twitter & \cmark & \xmark & \xmark & \cmark & \cmark \\
        CACT  & Clinical & \xmark & \cmark & \xmark & \tildemark & \cmark \\
        ExcavatorCovid & News & \xmark & \cmark & \cmark & \cmark & \xmark \\ 
        BioCaster & News & \xmark & \xmark & \cmark & \cmark & \xmark \\
        DANIEL & News & \xmark & \tildemark & \xmark & \tildemark & \cmark \\
        \bottomrule
    \end{tabular}
    \caption{Objective comparison of various epidemiological datasets COVIDKB \cite{zong-etal-2022-extracting}, CACT \cite{DBLP:journals/jbi/LybargerOTY21}, ExcavatorCovid \cite{min-etal-2021-excavatorcovid}, BioCaster \cite{DBLP:journals/bioinformatics/CollierDKGCTNDKTST08}, and DANIEL \cite{DBLP:journals/artmed/LejeuneBDL15} with our dataset \dataName{}. We objectify the source of data (Data Source), the level of annotation granularity (Sentence Level), the presence of trigger information (Trigger Present), the presence of social and personal events (Social Events and Personal Events), and the suitability of ontology for social media (SMG -- Social Media Granular). \tildemark{} indicates partial presence.}
    \label{tab:dataset-checklist}
\end{table}

\paragraph{Event Extraction Datasets}
\label{sec:ed-datasets}
Event Extraction (EE) is the task of detecting events (Event Detection) and extracting structured information about specific roles linked to the event (Event Argument Extraction) from natural text.
Earliest works for this task can be dated back to MUC \cite{sundheim-1992-overview, grishman-sundheim-1996-message} and the more standard ACE \cite{doddington-etal-2004-automatic}.
Over the years, ACE was extended to various datasets like ERE \cite{song-etal-2015-light} and TAC KBP \cite{DBLP:conf/tac/EllisGFKSBS15}.
Recent progress has been the creation of massive datasets and huge event ontologies with datasets like MAVEN \cite{wang-etal-2020-maven}, RAMS \cite{ebner-etal-2020-multi}, WikiEvents \cite{li-etal-2021-document}, DocEE \cite{tong-etal-2022-docee}, GENEVA \cite{parekh-etal-2023-geneva} and GLEN \cite{DBLP:journals/corr/abs-2303-09093}.
These ontologies and datasets cater to general-purpose events and do not comprise epidemiological event types.

\paragraph{Epidemiological Ontologies}
\label{sec:epi-ontologies}
Earliest works \cite{lindberg1993unified, rector1996galen} defined highly rich taxonomies for describing technical concepts used by biomedical experts.
Further developments led to the creation of SNOMED CT \cite{DBLP:conf/amia/StearnsPSW01} and PHSkb \cite{DBLP:journals/midm/DoyleMGH05} that define a list of reportable events used for communication between public health experts.
BioCaster \cite{DBLP:journals/bioinformatics/CollierDKGCTNDKTST08} and PULS \cite{DBLP:conf/tsd/DuEKNTY11} extended ontologies for the news domain.
Recent works of NCBI \cite{DBLP:journals/jbi/DoganLL14}, IDO \cite{DBLP:journals/biomedsem/BabcockBCS21} and DO \cite{DBLP:journals/nar/SchrimlMSOMFBJB22} focus on comprehensively organizing human diseases. 
In light of the recent COVID-19 pandemic, CIDO \cite{DBLP:conf/icbo/He0OWLHHBLDAXHY20} define a technical taxonomy for coronavirus, while ExcavatorCovid \cite{min-etal-2021-excavatorcovid} automatically extract COVID-19 events and relations between them.
Most of these ontologies are too fine-grained or limited to specific events, and can't be directly used for ED from social media, as also shown in Table~\ref{tab:dataset-checklist}.

\paragraph{Epidemiological Information Extraction}
\label{sec:epi-ie-works}
Early works utilized search-engine queries and click-through rates for predicting influenza trends \cite{DBLP:conf/amia/Eysenbach06, ginsberg2009detecting}.
Information extraction from Twitter has also been quite successful for predicting influenza trends \cite{signorini2011use, lamb-etal-2013-separating, paul2014twitter}.
Over the years, various biomedical monitoring systems have been developed like BioCaster \cite{DBLP:journals/bioinformatics/CollierDKGCTNDKTST08, DBLP:journals/bioinformatics/MengOPSPFGZKBC22}, HeathMap \cite{DBLP:journals/jamia/FreifeldMRB08}, DANIEL \cite{DBLP:journals/artmed/LejeuneBDL15}, EpiCore \cite{olsen2017epicore}.
Extensions to support multilingual systems has also been explored \cite{DBLP:journals/artmed/LejeuneBDL15, mutuvi-etal-2020-dataset, sahnoun-lejeune-2021-multilingual}.
For the COVID-19 pandemic, several frameworks like CACT \cite{DBLP:journals/jbi/LybargerOTY21} and COVIDKB \cite{zong-etal-2022-extracting, mendes-etal-2023-human} were developed for extracting symptoms, infection and treatment diagnosis.
Most of these systems are disease-specific, focus on news and clinical domains, and use keyword/rule-based or simple BERT-based models, as shown in Table~\ref{tab:dataset-checklist}.
In our work, we explore exploiting ED while focusing specifically on the social media domain.

\section{Conclusion and Future Work}

In this work, we develop an Event Detection (ED) framework to extract events from social media to provide early epidemic warnings.
To facilitate this, we create our Twitter-based dataset \dataName{} comprising seven event types.
Through experimentation, we show how existing models fail; while models trained on \dataName{} can effectively extract events and provide early warnings for unseen emerging epidemics.
More broadly, our work demonstrates how event extraction can exploit social media to aid policy-making for better epidemic preparedness.


\section*{Acknowledgements}
We thank I-Hung Hsu, Jiao Sun, Alex Spangher, Yu Zhou, Hritik Bansal, Yufei Tian, Chujie Zheng, and Rohan Wadhawan for their valuable insights, paper reviews, and constructive comments.
We thank the anonymous reviewers and the area editors across multiple rounds of reviews for their feedback.
This work was partially supported by NSF 2106859, 2200274, AFOSR MURI via Grant \#FA9550- 22-1-0380, Defense Advanced Research Project Agency (DARPA) grant \#HR00112290103/HR0011260656, and a Cisco Sponsored Research Award.

\section*{Limitations}

Our work focuses majorly on a single source of social media - Twitter.
We haven't explored other social media platforms and how ED would work on those platforms in our work.
We leave that for future work, but are optimistic that our models should be able to generalize across platforms.
Secondly, our work mainly only focuses on ED as the primary task, while its sister task Event Argument Extraction (EAE) is not explored.
We hope to extend our work for EAE as part of our future work.
Finally, we would like to show the generalization of our models on a vast range of diseases.
However owing to budget constraints and the lack of publically available Twitter data for other diseases, we couldn't perform such a study.
However, we believe showing results on three diseases lays the foundation for generalizability of our model.

\section*{Ethical Considerations}

One strong assumption in our work is the availability of internet and social media for discussions about epidemics.
Since not everyone has equal access to these platforms, our dataset, models, and results do not represent the whole world uniformly.
Thus, our work can be biased and should be considered with other sources for better representation.

Our dataset \dataName{} is based on actual tweets posted by people all over the world.
We attempted our best to anonymize any kind of private information in the tweets, but we can never be completely thorough, and there might be some private information embedded still in our dataset.
Furthermore, these tweets were sentimental and may possess stark emotional, racial, and political viewpoints and biases.
We do not attempt to clean any of such extreme data in our work (as our focus was on ED only) and these biases should be considered if being used for other applications.

Since our ED models are trained on \dataName{}, they may possess some of the dataset-based social biases.
Since we don't focus on bias mitigation, these models should be used with due consideration.

Lastly, we do not claim that our models can be used off-the-shelf for epidemic prediction as it hasn't been thoroughly tested and can have false positives and negatives too.
Furthermore, our results are shown on academic datasets and do not utilize all possible Twitter data.
We majorly throw light to show these model capabilities and motivate future work in this direction.
The usage of these systems for practical purposes should be appropriately considered.

\bibliography{anthology,custom}

\pagebreak
\clearpage

\appendix

\section{Ontology Creation - Additional Details}
\label{sec:ontology-creation-appendix}

\subsection{Complete ontology}
\label{sec:complete-ontology}

Here, we first describe the selection steps for event types for our ontology as follows:
\begin{enumerate}
    \item \textit{Curation of event types}: We scan through existing medical ontologies like BCEO \cite{DBLP:journals/bioinformatics/CollierDKGCTNDKTST08}, IDO \cite{DBLP:journals/biomedsem/BabcockBCS21}, and the ExcavatorCovid \cite{DBLP:conf/emnlp/MinRQZXM21} and curate a large list of event types for infectious and epidemic-related diseases.
    \item \textit{Merge event types across ontologies}: Since these existing ontologies may have repetitive event types, we perform a merging step. Specifically, two human experts manually examine and merge event types that are exactly similar in our curated list of event types.
    \item \textit{Filter out disease-specific event types}: Some event types in our curated list are specific to certain diseases. We identify and filter out such event types (e.g. Mask Wearing for COVID-19 which may not be observed for other diseases). We utilize opinions from public health experts to aid this step ensuring our event types are disease-agnostic.
    \item \textit{Definition Correction}: Utilizing aid from public health experts, we add and refine definitions for the curated set of event types and ensure they are disease-agnostic.
    \item Organization - Following ExcavatorCovid \cite{DBLP:conf/emnlp/MinRQZXM21}, we organize our curated list of event types into three larger categories: social (events involving larger populations), personal (individual-oriented events), and medical (medically focused events) types.
\end{enumerate}

Our complete initial event ontology comprises $18$ event types along with their event definitions organized into three abstract categories as shown in Table~\ref{tab:complete-ontology}.

\begin{table*}[h]
    \centering
    \small
    \resizebox{.99\textwidth}{!}{
    \begin{tabular}{p{1.8cm}|p{8.5cm}|p{3.5cm}}
        \toprule
        \textbf{Event name} & \textbf{Event Definition} &
        \textbf{Action for Final Ontology}\\
        \midrule
        \multicolumn{3}{c}{\textsc{Social Scale Events}}\\
        \midrule
        Prefigure & The signal that precedes the occurrence of a potential epidemic. & Discarded\\
        Outbreak & The process of disease spreading among a certain amount of the population at a massive scale. & Merged into \textit{Spread} \\
        Spread & The process of disease spreading among a certain amount of the population but at a local scale. & Final Event\\
        Control & Collective efforts trying to impede the spread of a epidemic. & Final Event\\
        Promote & The relationship of a disease driver leading to the breakout of a disease. & Discarded\\
        \midrule
        \multicolumn{3}{c}{\textsc{Personal Scale Events}}\\
        \midrule
        Prevent & Individuals trying to prevent the infection of disease. & Final Event\\
        Infect & The process of a disease/pathogen invading host(s). & Final Event\\
        Symptom & Individuals displaying physiological features indicating the abnormality of organisms. & Final Event\\
        Treatment & The process that a patient is going through with the aim of recovering from symptoms. & Merged into \textit{Cure}\\
        Cure & Stopping infection and relieving individuals from infections/symptoms. & Final Event \\
        Immunize & The process by which an organism gains immunization against an infectious agent. & Merged into \textit{Prevent}\\
        Death & End of life of individuals due to infectious disease. & Final Event\\
        \midrule
        \multicolumn{3}{c}{\textsc{Medical Scale Events}}\\
        \midrule
        Cause & The causal relationship of a pathogen and a disease. & Discarded\\
        Variant & An alternation of a disease with genetic code-carrying mutations. & Discarded\\
        Intrude & The process of an infectious agent intruding on its host. & Merged into \textit{Infect}\\
        Respond & The process of a host responding to an infection. & Discarded\\
        Regulate & The process of suppressing and slowing down the infection of a virus. & Merged into \textit{Cure}\\
        Transmission route & The process of a pathogen entering another host from a source. & Discarded\\
        \bottomrule
    \end{tabular}}
    \caption{Complete initial epidemic event ontology comprising $18$ event types along with their event definitions organized into $3$ higher-level abstract categories. We also present details about the action taken for each event type in the final ontology.}
    \label{tab:complete-ontology}
\end{table*}

\subsection{Initial analysis of events}
\label{sec:initial-event-analysis}

Our initial ontology (\S~\ref{sec:complete-ontology}) was constructed using previous ontologies and human knowledge.
But the relevance of each event type for social media (specifically Twitter) remains unknown.
To evaluate this relevance, we first associate each event type with event-specific keywords.
Then we utilize frequency and specificity as two guiding heuristics for further filtering/merging of event types in our curated ontology.
We utilize the base Twitter dataset for \dataName{} for conducting this analysis.
We describe each of these steps in more detail here:

\paragraph{Keyword Association}
In order to objectively conduct this analysis, we associate each event type with a set of keywords.\footnote{We release these keywords as part of our final code.}
This association involves two simple steps:
\begin{enumerate}
    \item \textit{Human expert curation}: For each event type, a human expert curates 2-3 simple yet specific keywords for each event based on commonsense knowledge. For example, for the \textit{Cure} event, the set of curated keywords were [cure, recovery].
    \item \textit{Thesaurus-based expansion}: For each human-expert curated list, we utilize an external resource - Thesaurus\footnote{\url{https://www.thesaurus.com/}} to further find event-relevant keywords. Human experts manually curate keywords from this thesaurus list such that the curated keyword is not generic (e.g. \textit{display} is filtered out for event \textit{Symptom} since it has other meanings as well).
\end{enumerate}

We provide some example keywords for each event type along with some social media posts in which they appear in Table~\ref{tab:event-keywords-posts}.
We can also see how some example social media posts with the keywords can have different meanings as well. But we ignore them as this is a preliminary analysis.

\begin{table*}[h]
    \centering
    \small
    \resizebox{.99\textwidth}{!}{
    \begin{tabular}{p{1.8cm}|p{3cm}|p{8.5cm}}
        \toprule
        \textbf{Event name} & \textbf{Associated Keywords} &
        \textbf{Example Social Media Post}\\
        \midrule
        \multicolumn{3}{c}{\textsc{Social Scale Events}}\\
        \midrule
        Prefigure & alert, foreshadow, warn, indicate, ...  & (user) Alabama hospitals with no ICU beds \textbf{foreshadow} rural coronavirus crisis - Business Insider (url)\\ \hline
        Outbreak & epidemic, pandemic, outbreak, ... & (user) If Boris goes, who is Nicola going to follow in the coronavirus \textbf{epidemic}? \\ \hline
        Spread & outbreak, spread, contagious, ... & COVID-19 \textbf{outbreak} detected in immigration detention centres ... \\ \hline
        Control & quarantine, isolate, restriction, protocol, ... & The doctors in my family are losing their minds seeing these videos of people flagrantly violating social distancing \textbf{protocols}. \\ \hline
        Promote & boost, lead, accelerate & (user) Covid-19 Will \textbf{Accelerate} the AI Health Care Revolution (url) \\
        \midrule
        \multicolumn{3}{c}{\textsc{Personal Scale Events}}\\
        \midrule
        Prevent & prevent, protect, avoid, vaccine, ... &  (user) why can’t they wear masks and god will cover them anyways then you have double the \textbf{protection} \\ \hline
        Infect & infect, spread, outbreak, infection, ... & (user) warns of COVID-19 \textbf{spread} in group gatherings after 2 people infect dozens at church. (url) \\ \hline
        Symptom & symptom, fever, ill, sick, ache, ... & ... average person dying from coronavirus has serious \textbf{illness} and is about to die anyway. \\ \hline
        Treatment & doctor, drug, hospitalize, nurse, ... & Hyperbaric oxygen \textbf{therapy} can help fight one of the major issues(lack of oxygen) caused by covid/corona virus?  \\ \hline
        Cure & recover, cure, heal, ... & Patients who've \textbf{recovered} from Covid-19 and test positive again aren't contagious, a study says (url) \\ \hline
        Immunize & immunize, antibody, vaccine, ... & We need to hope the number of cases go up in order to Herd \textbf{immunize}, and continually watch the \# of hospitalizations. \\ \hline
        Death & death, decease, kill, ... & Highest \textbf{death} toll in Europe. Economy on the floor. Quarantine for international travellers months after the horse bolted. \\
        \midrule
        \multicolumn{3}{c}{\textsc{Medical Scale Events}}\\
        \midrule
        Cause & cause, lead & Once again, 1-2 infections \textbf{lead} to a large number of new infections. \\ \hline
        Variant & lineage, variant, mutate & \#DeltaVariant surging in U.S. New data show Delta much more contagious than previous \textbf{versions} of \#COVID19.\\ \hline
        Intrude & attack, intrude, harm & (user) Peru found 100\% cure zero \textbf{harm} ivermectin. Pinned tweet medical paper. Plagues over. Retweet. \\ \hline
        Respond & respond, react & Aid agencies face huge challenges \textbf{responding} to \#COVID19 in countries affected by conflict ...  \\ \hline
        Regulate & regulate, limit, decelerate, slow, ... & How can you \textbf{regulate} a child in school after lockdown? \\ \hline
        Transmission route & airborne, contact, air, water, ... & (user) Very unfortunate. You can't workout with a mask on, is Corona \textbf{airborne}? Just asking.. \\
        \bottomrule
    \end{tabular}}
    \caption{All the events in our ontology along with their associated keywords and example social media posts.}
    \label{tab:event-keywords-posts}
\end{table*}

\begin{figure*}[t]
    \centering
    \includegraphics[width=13cm]{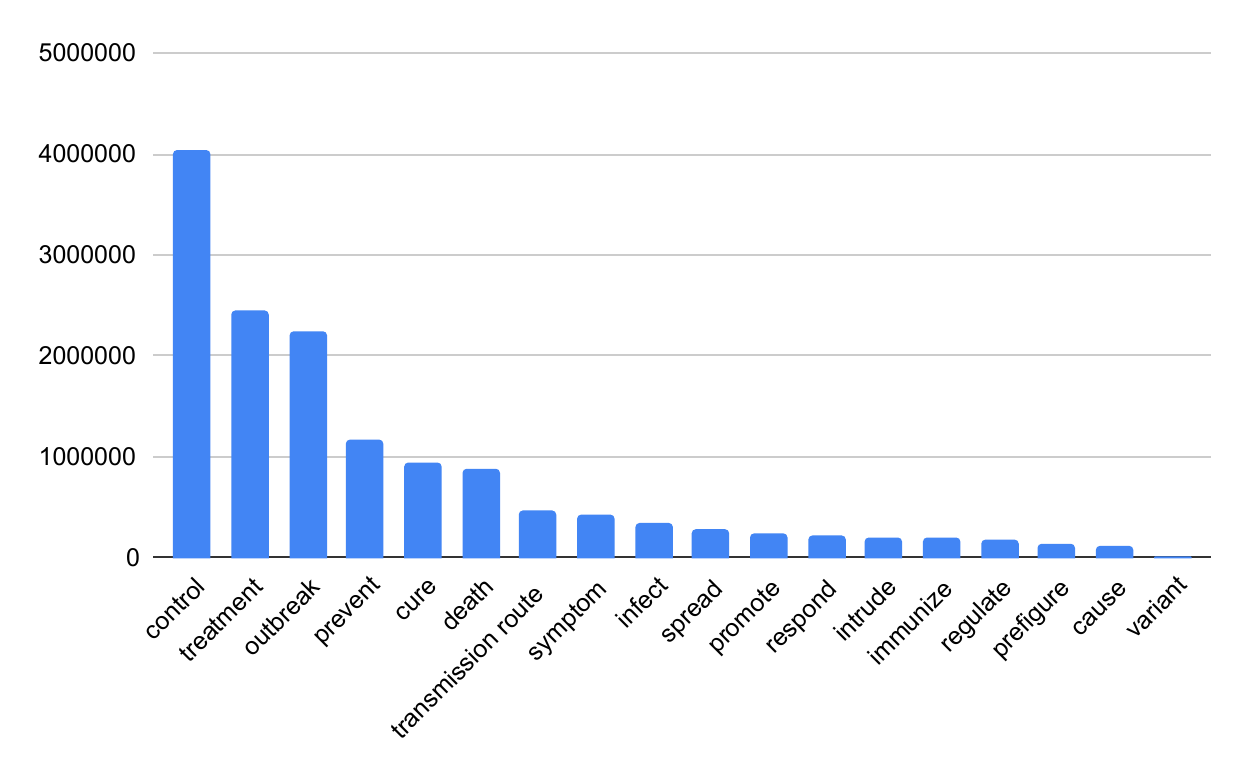}
    \caption{Frequency of occurrence based on keyword search for all event types in the initial complete ontology.}
    \label{fig:initial-event-analysis}
\end{figure*}

\paragraph{Frequency-based filtering}
Using frequency, we aim to filter out event types that are less mentioned in social media.
To approximately estimate the frequency of each event type in social media, we count the number of social media posts containing any of the curated keywords for each event type.
We show the keyword-count based frequency for each event type in Figure~\ref{fig:initial-event-analysis}.
We observe that most events under the medical abstraction occur much lesser than others.
Furthermore, the variance in frequency is large as the most frequent event type \textit{control} is 180 times more likely to occur than the least frequent event type \textit{variant}.
Since such low-frequency events (e.g. \textit{Variant}, \textit{Cause}, \textit{Prefigure}, etc.) are less likely to be mentioned in a smaller sample of data, we discard or merge such events for our final ontology.

\paragraph{Specificity-based filtering}
Specificity ensures that each event type is uniquely identifiable with a good confidence and mainly aims to reduce ambiguity and make the event types more distinct.
To estimate specificity, for each curated keyword of an event type, we randomly sample a small number of non-duplicate social media posts.
Human experts then manually evaluate the keyword specificity based on the percentage of posts wherein the semantic meaning of the keyword matches the definition of its event and is specific only to this event type.
This specificity and distinctivity classifies keywords as high, medium, or low.

For example, the \textit{Control} event comprises high specificity keywords such as \textit{quarantine}, \textit{protocol}, \textit{guidelines}; medium specificity keywords such as \textit{restrict}, \textit{postpone}, \textit{investigate}; and low specificity keywords such as \textit{battle}, \textit{separation}, \textit{limitation}.
On the other hand, the event \textit{Prefigure} doesn't have any high specificity keywords, but only medium specificity keywords such as \textit{foreshadow} and low specificity keywords such as \textit{foretell}.

Our analysis suggests that medium and low specificity keywords are more likely to give false positives relative to high specificity ones.
Thus, we filter/merge event types that have a high number of low-confidence keywords (e.g. \textit{Intrude}, \textit{Promote}).

\paragraph{Final Ontology}
Thus, with the above filtering and merging, we shrink our ontology from $18$ event types to seven event types that are distinguishable, frequent, and have a low false-positive rate.
We provide details about the action taken for each event type with respect to the final ontology in Table~\ref{tab:complete-ontology}.

\subsection{Association of Seed Tweets with Events}
\label{sec:seed-tweets}

For event-based filtering, we associate each event type with a set of 5-10 seed tweets.
In Table~\ref{tab:seed-tweets}, we provide a couple of the seed tweets for each event type for reference.
We also release these seed tweets as part of the main code.

\begin{table}[h]
    \centering
    \small
    \begin{tabular}{l|p{5.6cm}}
        \toprule
        \textbf{Event} & \textbf{Sample Seed} \\
        \textbf{Name} & \textbf{Tweets} \\
        \midrule
        Infect & The pandemic will infect many older people \\
        & My brother tested positive for COVID-19\\
        \midrule
        Spread & The COVID-19 has spread throughout Europe \\
        & Number of positive cases is rising\\
        \midrule
        Symptom & Fever is a key symptom of COVID-19 \\
        & I became incredibly sick after catching the COVID\\
        \midrule
        Prevent & Officials hope the vaccine will prevent the spread of COVID-19 in high-risk populations \\
        & Medical experts encourage young kids to wash their hands\\
        \midrule
        Control & Many countries are requiring mask wearing to reduce the spread of the pandemic\\
        & Government officials have imposed a lockdown on certain districts\\
        \midrule
        Cure & There is no effective treatment for COVID so far\\
        & Even patients already recovered from COVID remain coughing for a while\\
        \midrule
        Death & My friend has passed away because of the pandemic\\
        & Millions of people are dead because of COVID\\
        \bottomrule
    \end{tabular}
    \caption{Sample seed tweets for the different event types in our ontology.}
    \label{tab:seed-tweets}
\end{table}

\subsection{Coverage analysis of ontology}
\label{sec:disease-coverage}
To quantitatively verify the coverage of our ontology, we conduct an analysis on four diseases with very different characteristics - COVID-19, Monkeypox, Dengue, and Zika.
For each disease, we randomly sample $300$ tweets and then filter them if they are related to the disease or not.
Next, we annotate the filtered disease-related tweets based on our ontology and 
evaluate the proportion of event occurrences relative to the number of disease-related tweets.
We find that our ontology has high coverage of $50\%$ for COVID-19, $44\%$ for Monkeypox, $70\%$ for Dengue, and $73\%$ for Zika.
This in turn assures that our ontology can be used to detect epidemic events for various different kinds of diseases.

\begin{figure}[h]
    \centering
    \includegraphics[width=\columnwidth]{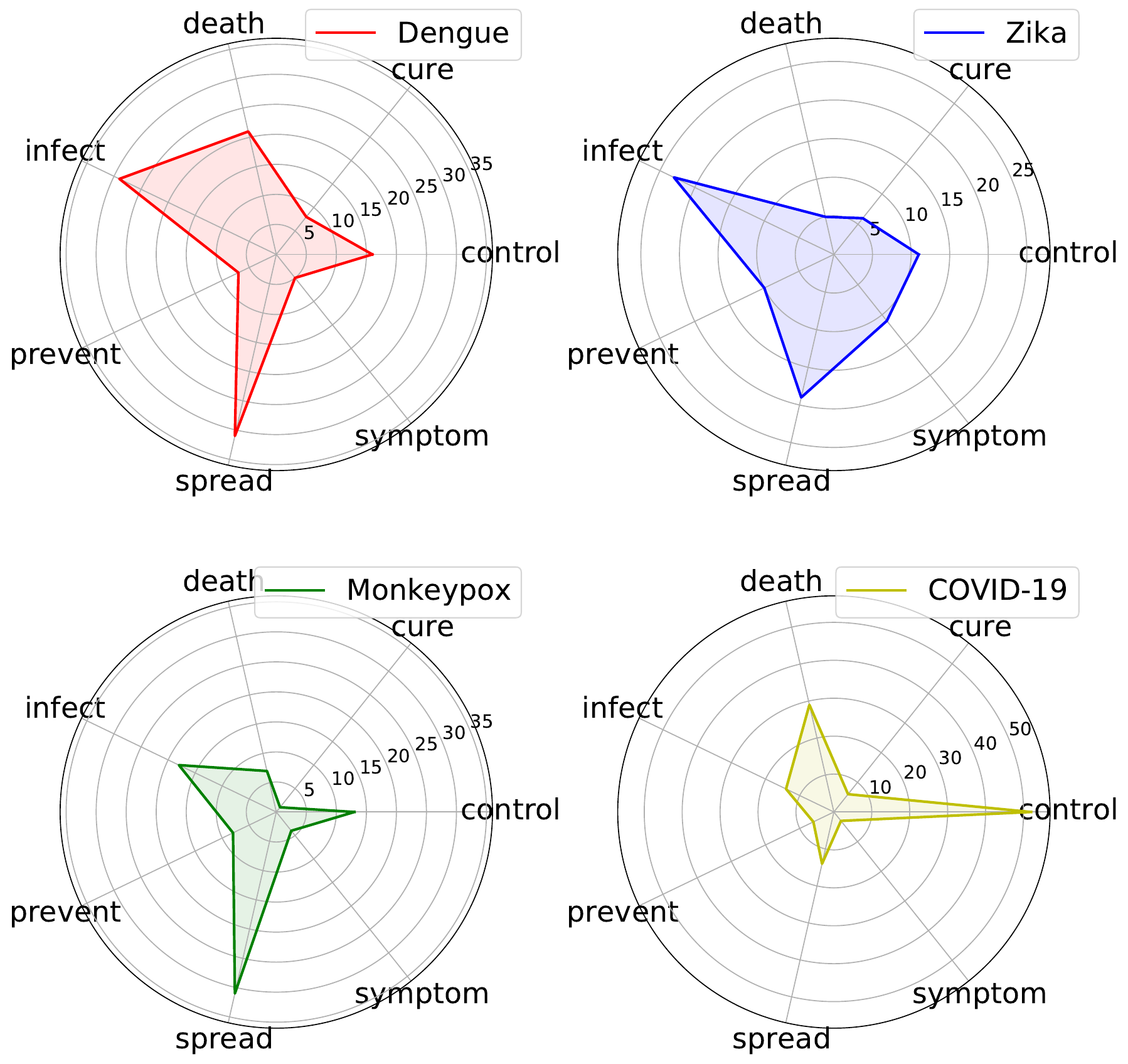}
    \caption{Event type distribution of the disease-related tweets for each disease. Numbers on the axis represent count of mentions for a given event type.}
    \label{fig:disease-specific-event-type-distribution}
\end{figure}

\paragraph{Event Type Distribution}
As part of our analysis, we also study our ontology's event type distribution for each disease and its correlation with the disease properties and outbreak stage.
We show this event distribution in Figure~\ref{fig:disease-specific-event-type-distribution} for each of the diseases.
We note that distributions for Dengue and Monkeypox exhibit a strong focus on \textit{spread} and \textit{infect} events.
This makes sense as the data for these diseases was collected at earlier stages of the outbreak when mitigation measures were not being discussed yet.
On the other hand, for COVID-19, the distribution is vastly dominated by \textit{control} and \textit{death} events.
Our COVID-19 data was collected in May 2020 when the outbreak had vastly spread in America.
Thus our distribution reflects more notions of lockdowns and control measures as well reflects the deadly nature of the disease.




\section{Uniform Sampling v/s Random Sampling for Data Selection}
\label{sec:uniform-sampling}

Previously \citet{parekh-etal-2023-geneva} had shown how uniform sampling of data for events can yield more robust model performance.
To validate the same for our ontology and data, we conduct additional experiments comparing uniform sampling with random sampling.
More specifically, we annotate $200$ tweets that conform to a `real distribution'\footnote{Event-based filtering was still applied before sampling.} based on random sampling and compare the trained models on this data with models trained on $200$ tweets of uniform-sampling data.
We further annotated $300$ tweets based on the `real-distribution' which was used for the evaluation of these two sampling techniques.

\begin{table}[h]
    \centering
    \small
    \begin{tabular}{lcc}
        \toprule
        \textbf{Model} & \textbf{Tri-I} & \textbf{Tri-C} \\
        \midrule
        \multicolumn{3}{c}{\small \textsc{Trained on Uniform Distribution}} \\
        \midrule
        BERT-QA & $\textbf{58.19}$ & $52.30$ \\
        DEGREE & $55.83$ & $\textbf{52.88}$ \\
        TagPrime & $55.48$ & $50.51$ \\
        DyGIE++ & $53.22$ & $47.64$ \\
        \hline
        \textbf{Average} & $55.68$ & $50.83$ \\
        \midrule
        \multicolumn{3}{c}{\small \textsc{Trained on Random Distribution}} \\
        \midrule
        BERT-QA & $46.11$ & $43.76$ \\
        DEGREE & $46.11$ & $45.23$ \\
        TagPrime & $25.03$ & $24.15$ \\
        DyGIE++ & $\textbf{51.10}$ & $\textbf{47.35}$ \\
        \hline
        \textbf{Average} & $42.09$ & $40.12$ \\
        \bottomrule
    \end{tabular}
    \caption{Benchmarking ED models trained on uniformly-sampled and randomly-sampled \dataName{} data on real-distribution based test data of $300$ samples.}
    \label{tab:real_uniform_training-results}
\end{table}

We present our results in Table~\ref{tab:real_uniform_training-results} averaged over three model runs.
We show that in terms of best model performance, uniform sampling is better by $5.5$ F1 points compared to random sampling.
On average, uniform-sampling trained models outperform the random-sampling trained models by up to $11$ points.
Both these results prove how despite train-test distribution differences, uniform sampling leads to better training of downstream models.


\paragraph{Generalizability to Other Diseases}
We also evaluate the models trained on the uniform and random-sampled data for generalizability to other diseases of Monkeypox, Zika, and Dengue.
We show the results in Table~\ref{tab:real_uniform_gen_results}.
Clearly, we can see superior generalizability of uniform-sampling trained models as they outperform random-sampling trained models by $37$ F1 points for Monkeypox and $28$ F1 points for Zika + Dengue.
Overall, this result strongly highlights the impact of uniform sampling for robust and generalizable model training.

\begin{table}[h]
    \centering
    \small
    \begin{tabular}{lcc|cc}
        \toprule
        \multirow{2}{*}{\textbf{Model}} & \multicolumn{2}{c|}{\textbf{Monkeypox}} & \multicolumn{2}{c}{\textbf{Zika + Dengue}} \\
        & \textbf{Tri-I} & \textbf{Tri-C} & \textbf{Tri-I} & \textbf{Tri-C}\\
        \midrule
        \multicolumn{5}{c}{\small \textsc{Trained on Uniform Sampled Data}} \\
        \midrule
        BERT-QA & $56.56$ & $49.30$ & $56.35$ & $46.19$ \\
        DEGREE & $58.35$ & $53.39$ & $\textbf{58.37}$ & $\textbf{51.27}$ \\
        TagPrime & $\textbf{58.36}$ & $\textbf{53.56}$ & $57.05$ & $48.53$ \\
        DyGIE++ & $55.73$ & $48.30$ & $56.90$ & $47.10$ \\
        \midrule
        \multicolumn{5}{c}{\small \textsc{Trained on Real Sampled Data}} \\
        \midrule
        BERT-QA & $9.48$ & $7.97$ & $21.68$ & $20.43$ \\
        DEGREE & $10.76$ & $10.53$ & $19.33$ & $19.00$ \\
        TagPrime & $10.37$ & $8.57$ & $12.78$ & $12.28$ \\
        DyGIE++ & $19.59$ & $16.62$ & $26.43$ & $23.40$ \\
        \bottomrule
    \end{tabular}
    \caption{Generalizability benchmarking of ED models trained on $200$ samples of uniformly-sampled and randomly-sampled COVID data on other diseases of Monkeypox, Zika, and Dengue.}
    \label{tab:real_uniform_gen_results}
\end{table}

\section{Annotation Guidelines and Interface}
\label{sec:annotation-guidelines-interface}


\subsection{Annotation Guidelines}
\label{sec:annotation-guidelines}

\begin{figure*}[h]
    \centering
    \includegraphics[width=15cm]{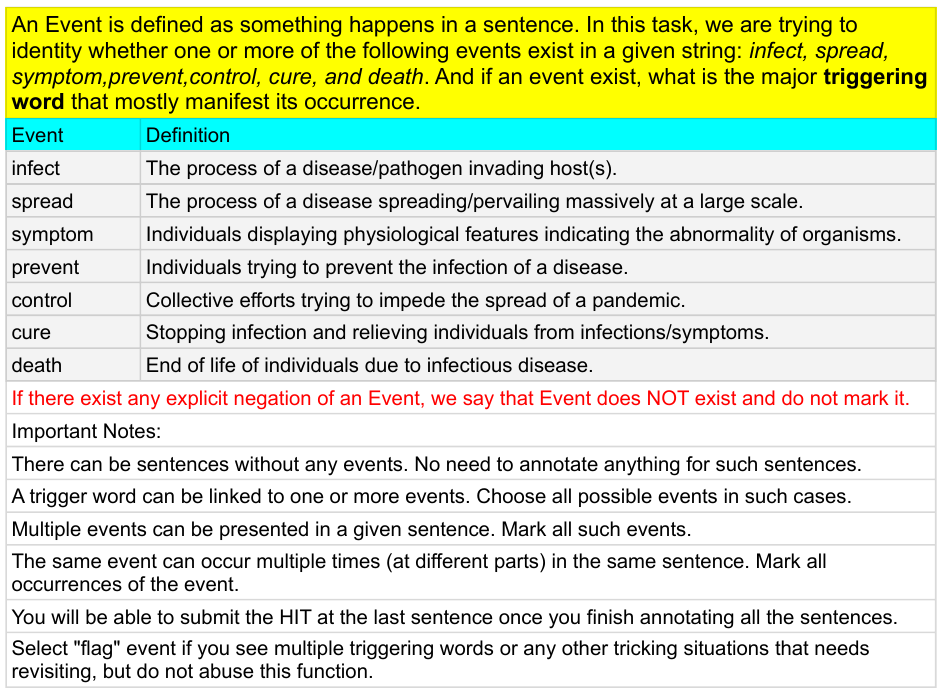}
    \caption{Task summary and the major annotation guidelines.}
    \label{fig:guidelines-task-summary}
\end{figure*}

\begin{figure*}[h]
    \centering
    \includegraphics[width=15cm]{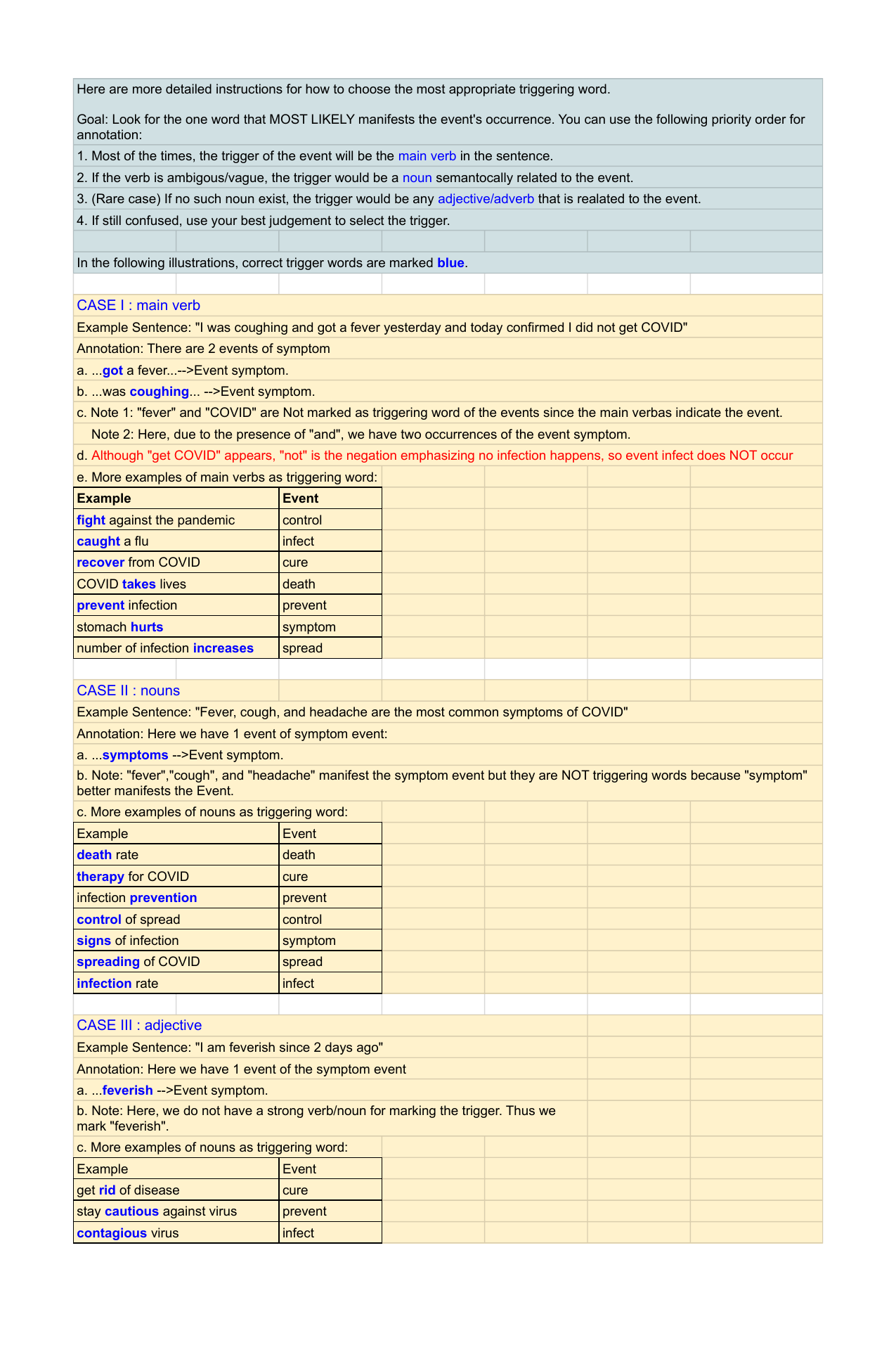}
    \caption{Guidelines to choose the proper triggering word.}
    \label{fig:guidelines-trigger-selection}
\end{figure*}

\begin{figure*}[h]
    \centering
    \includegraphics[width=15cm]{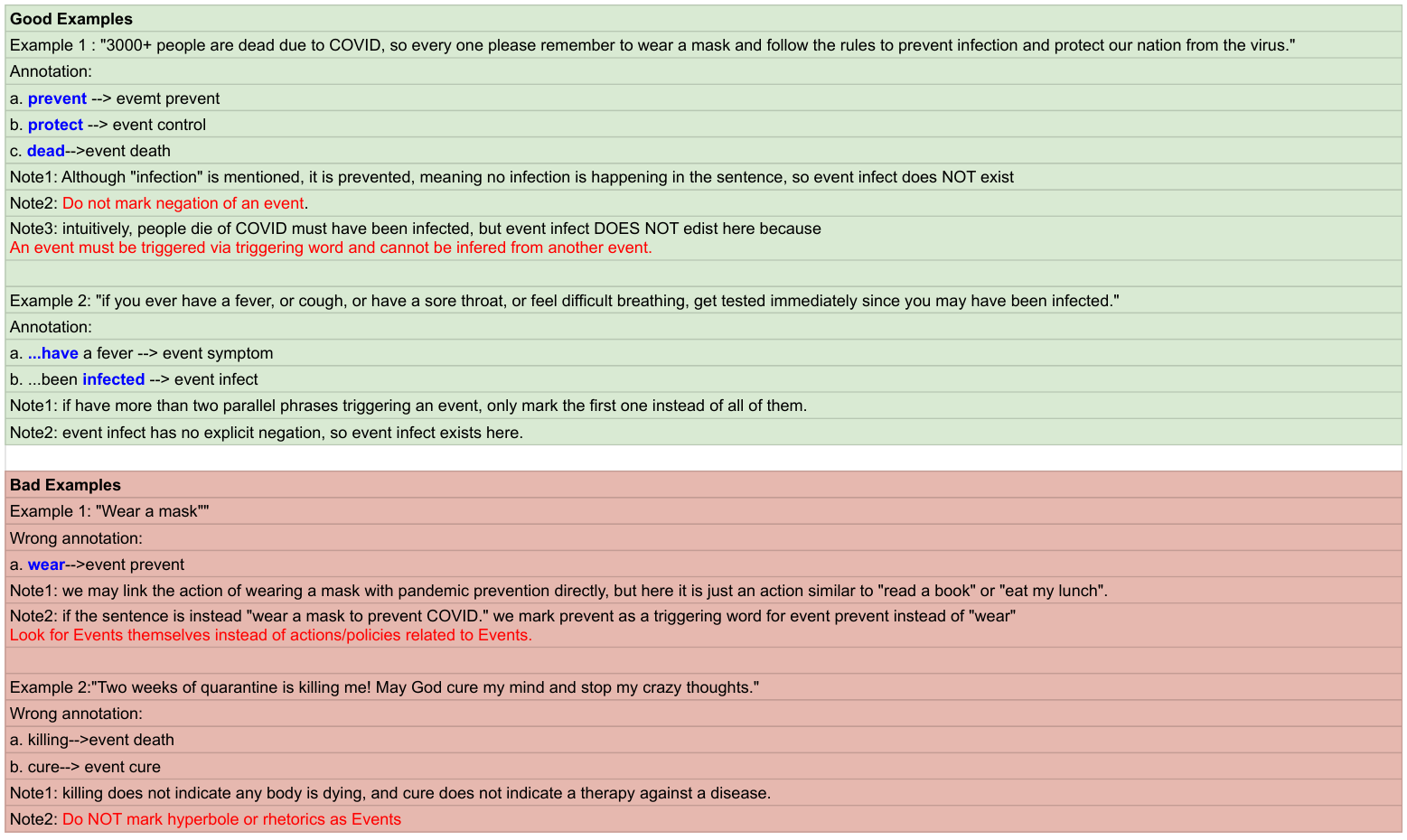}
    \caption{Positive and Negative examples in the annotation guideline.}
    \label{fig:guidelines-example-annotations}
\end{figure*}

Inspired by \citet{doddington-etal-2004-automatic}, we develop an extensive set of instructions with tricky cases and examples that have been developed through multiple rounds of expert annotation studies.
We present the task summary with the major instructions in Figure~\ref{fig:guidelines-task-summary}.
To reduce ambiguity in trigger selection, we present extensive examples and tricky cases with priority orders as shown in Figure~\ref{fig:guidelines-trigger-selection}.
Finally, we also provide a wide range of annotated positive and negative examples as part of the guidelines and show those in Figure~\ref{fig:guidelines-example-annotations}.

\subsection{Annotation Interface}
\label{sec:annotation-interface}

\begin{figure*}[h]
    \centering
    \includegraphics[width=15cm]{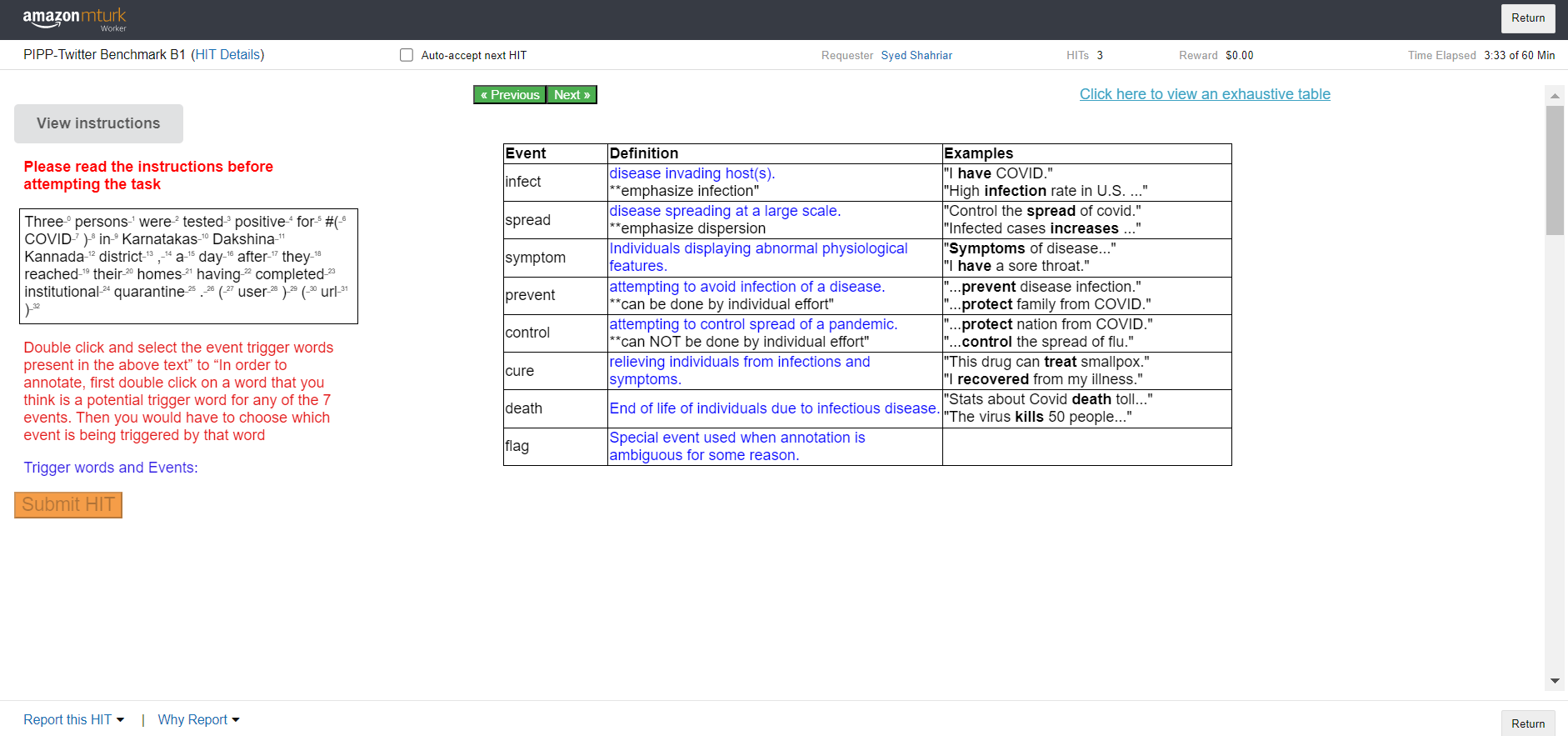}
    \caption{Illustration of the default annotation interface on Amazon Mechanical Turk.}
    \label{fig:interface-basic}
\end{figure*}

\begin{figure*}[h]
    \centering
    \includegraphics[width=15cm]{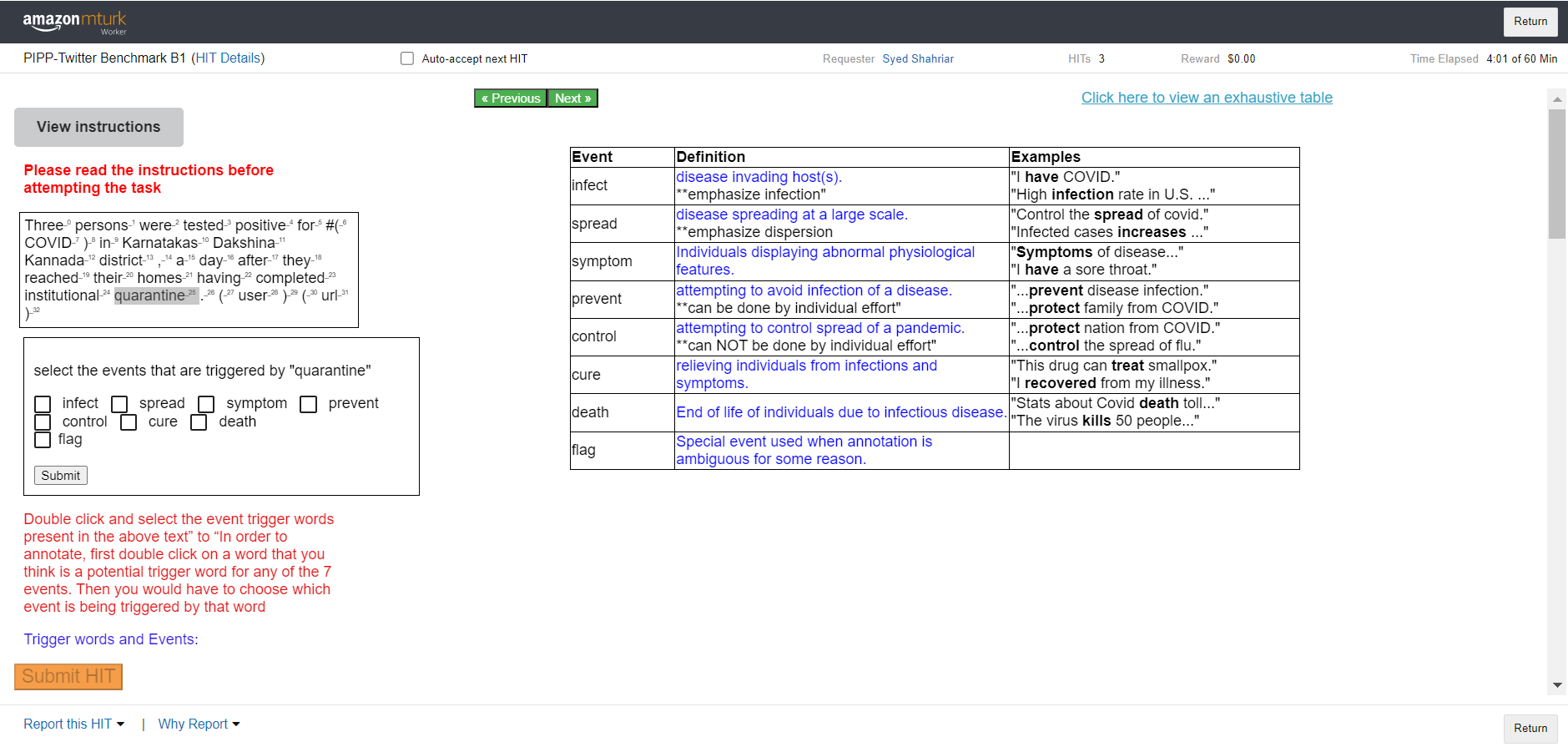}
    \caption{Illustration of selection of a word within a tweet for annotation in the interface.}
    \label{fig:interface-word-select}
\end{figure*}

\begin{figure*}[h]
    \centering
    \includegraphics[width=15cm]{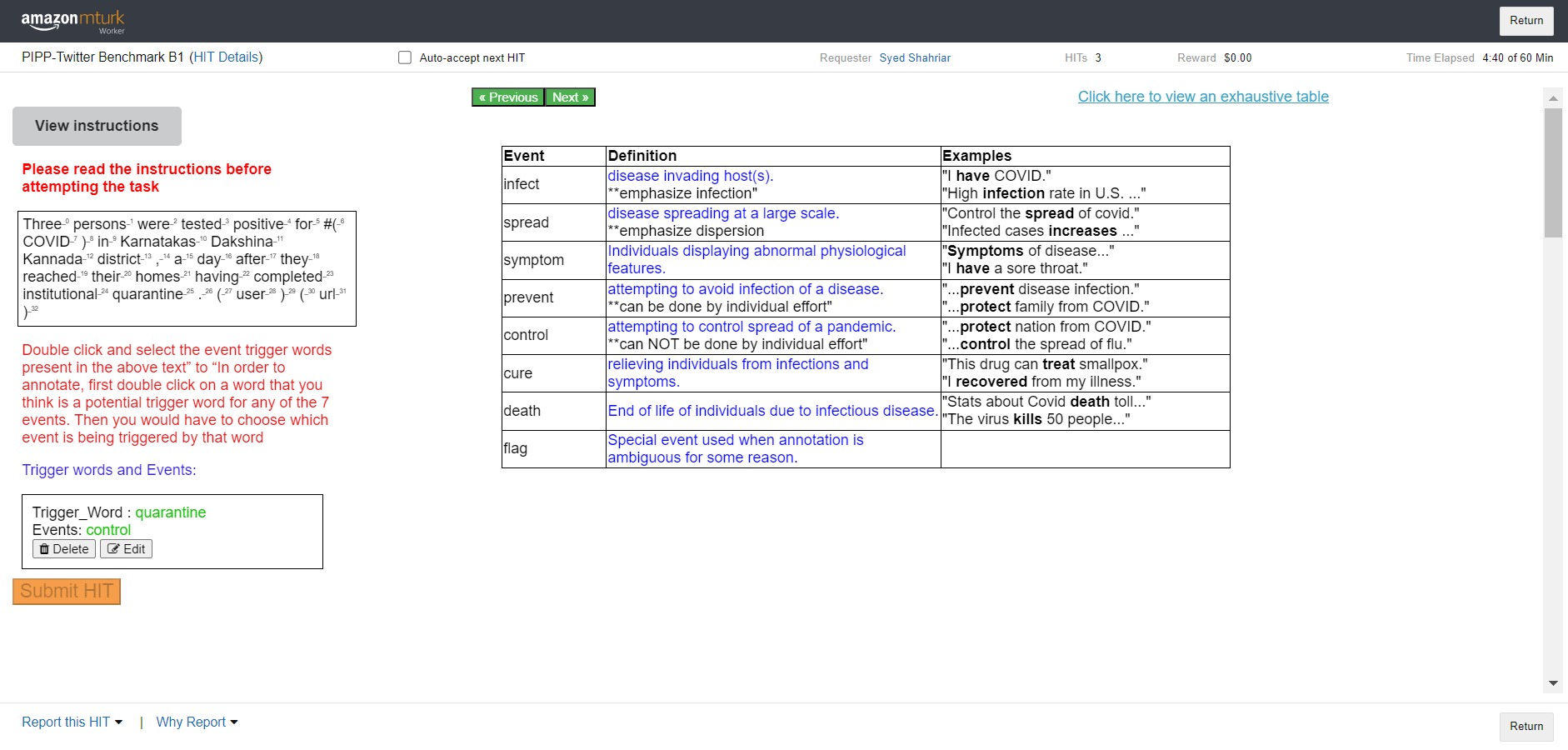}
    \caption{Illustration of the format and options available for a completed annotation in the interface.}
    \label{fig:interface-format-utilities}
\end{figure*}

We utilize Amazon Mechanical Turk\footnote{\url{https://www.mturk.com/}} as the interface for quick annotation.
To annotate, annotators can select any word and label it into one of the seven pre-defined event types.
Event definitions and examples are provided alongside for reference.
Each batch (also known as HIT) comprises five tweets for flexibility in annotations.
We show the interface and various utilities in Figure~\ref{fig:interface-basic}, ~\ref{fig:interface-word-select}, and ~\ref{fig:interface-format-utilities} respectively.

\section{Data Analysis for \dataName{}}
\label{sec:data-analysis-appendix}

\subsection{Event Coverage for previous datasets}
\label{sec:event-coverage}

To show the distinction of the event types covered in \dataName{} compared to other previous datasets, we calculate the percentage event types from \dataName{} present in various diverse previous dataset ontologies.
We show the results of this analysis in terms of partial coverage (similar events present) and exact coverage (exact event present) in Table~\ref{tab:dataset-coverage}.

\begin{table}[t]
    \centering
    \small
    \begin{tabular}{lrr}
        \toprule
        \multirow{2}{*}{\textbf{Dataset}} & \textbf{Partial} & \textbf{Exact} \\
        & \textbf{Match} & \textbf{Match} \\
        \midrule
        ACE \cite{doddington-etal-2004-automatic} & $14\%$ & $0\%$ \\
        ERE \cite{song-etal-2015-light} & $14\%$ & $0\%$ \\
        MAVEN \cite{wang-etal-2020-maven} & $42\%$ & $0\%$ \\
        MEE \cite{pbveyseh-etal-2022-mee} & $14\%$ & $0\%$ \\
        M$^2$E$^2$ \cite{li-etal-2020-cross} & $14\%$ & $0\%$ \\
        MLEE \cite{DBLP:journals/bioinformatics/PyysaloOMCTA12} & 0\% & 0\% \\
        FewEvent \cite{DBLP:conf/wsdm/DengZKZZC20} & 28\% & 0\% \\
        \bottomrule
    \end{tabular}
    \caption{Comparison of \dataName{} with ACE and MAVEN in terms of unique trigger words and average number of triggers per event mention. Avg = Average.}
    \label{tab:dataset-coverage}
\end{table}

Overall, from the table, we can note that there is no dataset with exact matches with our ontology.
This proves the distinctive coverage of our dataset.




\subsection{Event Distribution Analysis}
\label{sec:event-distribution-appendix}

\begin{figure}[h]
    \centering
    \includegraphics[width=\columnwidth]{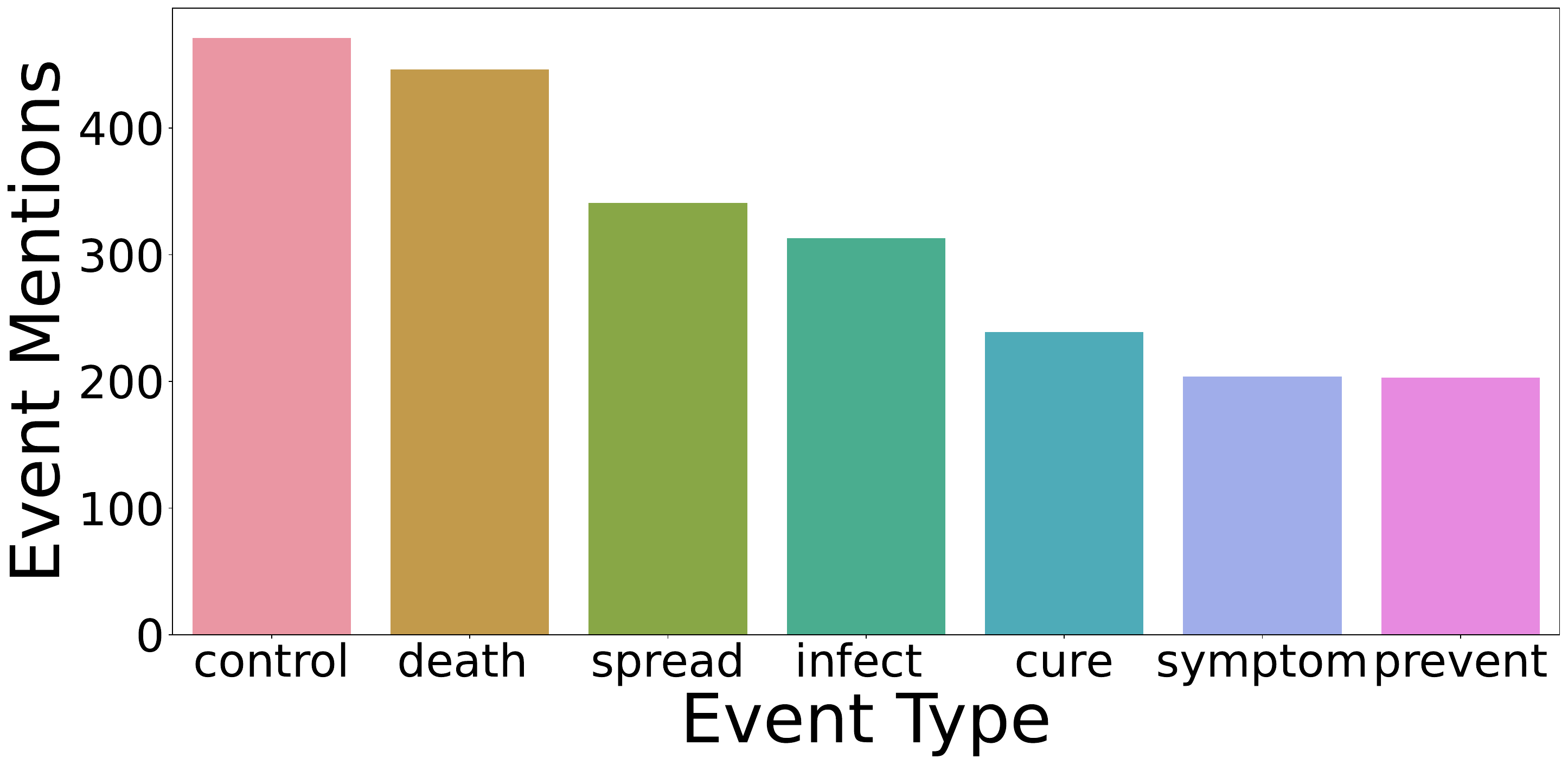}
    \caption{Distribution of event mentions per event type for our dataset \dataName.}
    \label{fig:event-type-distribution}
\end{figure}

\begin{figure}[h]
    \centering
    \includegraphics[width=\columnwidth]{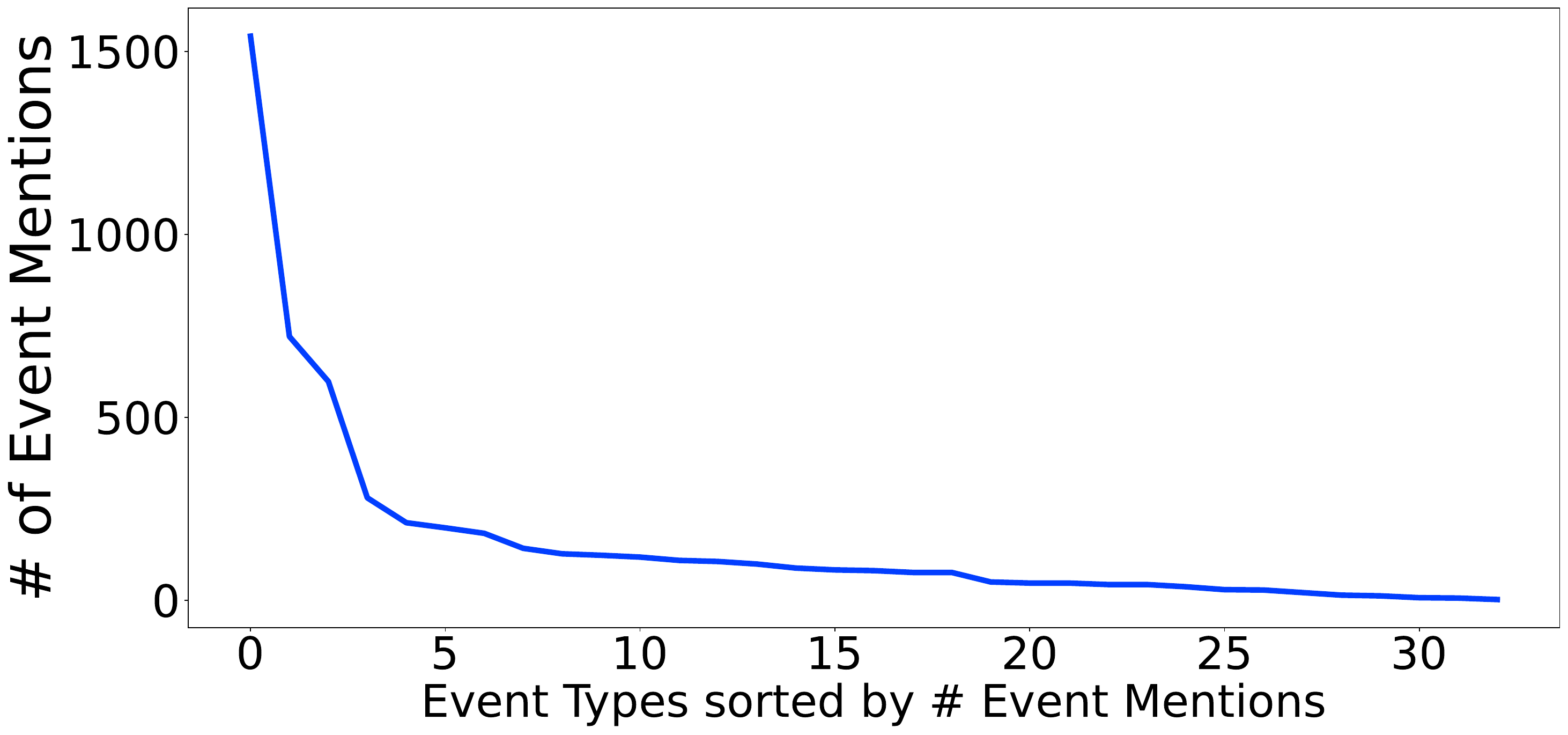}
    \caption{Distribution of event mentions for the event types in the ACE dataset.}
    \label{fig:ace-event-distribution}
\end{figure}

\begin{figure}[h]
    \centering
    \includegraphics[width=\columnwidth]{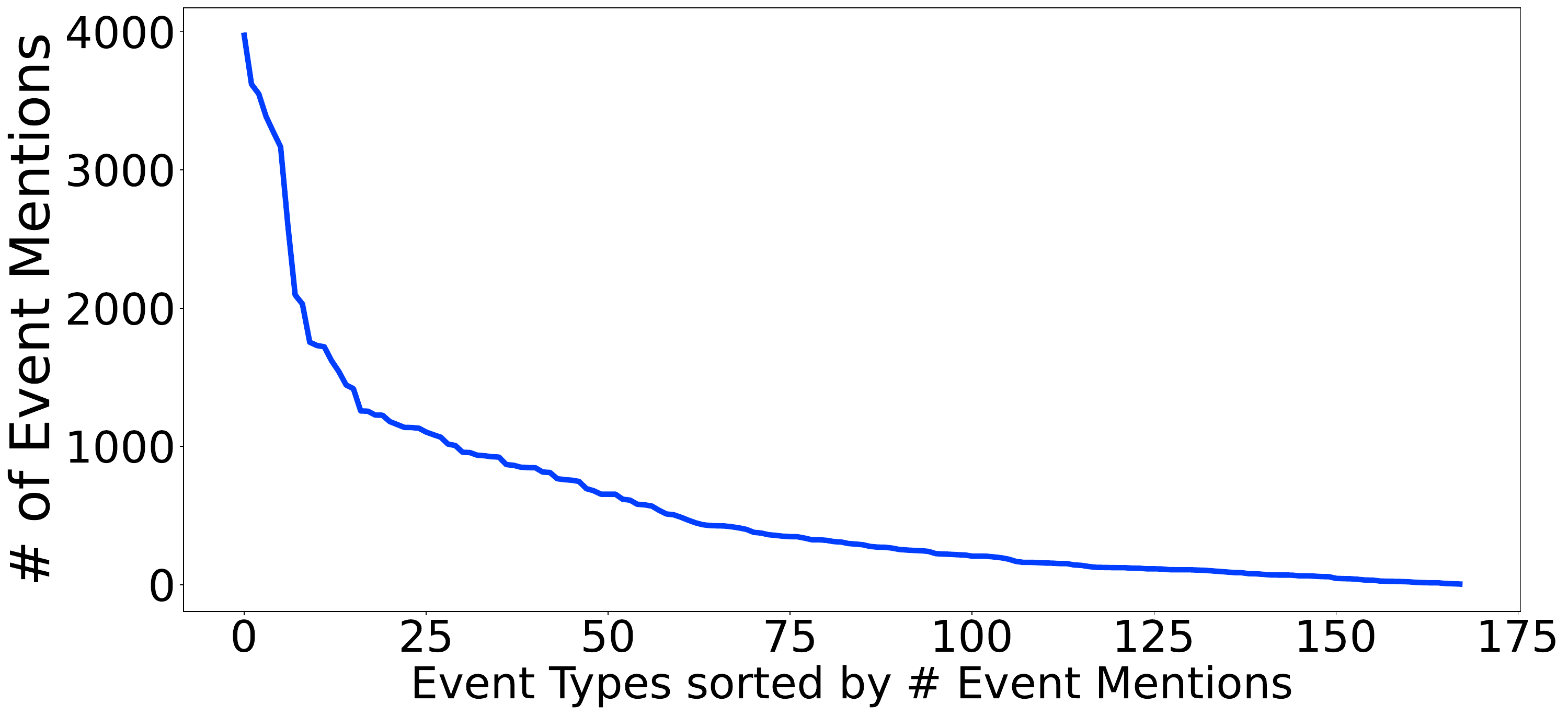}
    \caption{Distribution of event mentions for the event types in the MAVEN dataset.}
    \label{fig:maven-event-distribution}
\end{figure}

As part of data processing, we attempt to sample tweets in a more uniform distribution between the event types (\S~\ref{sec:twitter-selection}).
In Figure~\ref{fig:event-type-distribution}, we show the distribution of our dataset in terms of event types.
In contrast to tail-ending distributions of other standard datasets like ACE \cite{doddington-etal-2004-automatic} and MAVEN \cite{wang-etal-2020-maven} as shown in Figures~\ref{fig:ace-event-distribution} and ~\ref{fig:maven-event-distribution} respectively, our distribution of event mentions is more uniform.

\subsection{Monkeypox Test Data Statistics}
\label{sec:mpox-data-statistics}

We share the data statistics of the evaluation dataset used for Monkeypox in Table~\ref{tab:mpox-data-statistics} split according to each event type.
We observe that there is a disparity in distribution across different event types, with \textit{spread} mostly discussed and \textit{cure} and \textit{death} are least discussed.

\begin{table}[h]
    \centering
    \small
    \begin{tabular}{lr}
        \toprule
        \textbf{Event Type} & \textbf{\# Event Mentions} \\
        \midrule
        infect & $78$ \\
        spread & $119$ \\
        symptom & $43$ \\
        prevent & $70$ \\
        control & $62$ \\
        cure & $13$ \\
        death & $13$ \\
        \midrule
        \textbf{Total} & $\textbf{389}$ \\
        \bottomrule 
    \end{tabular}
    \caption{Data Statistics for the evaluation dataset for Monkeypox Event Detection categorized by event types.}
    \label{tab:mpox-data-statistics}
\end{table}

\subsection{Zika + Dengue Test Data Statistics}
\label{sec:zika-data-statistics}

We share the data statistics of the evaluation dataset used for Zika + Dengue in Table~\ref{tab:zika-data-statistics} split according to each event type.
We observe a more even distribution of event types with more focus on \textit{infect}, \textit{spread}, and \textit{death} well-discussed.

\begin{table}[h]
    \centering
    \small
    \begin{tabular}{lr}
        \toprule
        \textbf{Event Type} & \textbf{\# Event Mentions} \\
        \midrule
        infect & $57$ \\
        spread & $53$ \\
        symptom & $34$ \\
        prevent & $22$ \\
        control & $28$ \\
        cure & $20$ \\
        death & $60$ \\
        \midrule
        \textbf{Total} & $\textbf{274}$ \\
        \bottomrule 
    \end{tabular}
    \caption{Data Statistics for the evaluation dataset for Zika+Dengue Event Detection categorized by event types.}
    \label{tab:zika-data-statistics}
\end{table}

\section{ED models and Implementation Details}
\label{sec:implementation-details}

We present details about each ED model that we benchmark along with the extensive set of hyperparameters and other implementation details.

\subsection{TE}
TE \cite{lyu-etal-2021-zero} is a pre-trained model that formulates ED as a textual entailment and question-answering task.
We run our experiments for TE on an NVIDIA 1080Ti machine with support for 8 GPUs. Our hyperparameters are as listed in the original paper.

\subsection{WSD}

WSD \cite{yao-etal-2021-connect} is a classification model using on the joint encoding of the contextualized trigger and event definitions.
We run our experiments for WSD on an NVIDIA A100 machine with support for 8 GPUs. The major hyperparameters for this model are listed in Table~\ref{tab:hyper-wsd}.

\begin{table}[h]
    \centering
    \small
    \begin{tabular}{lr}
        \toprule
        Pre-trained LM & RoBERTa-Large \\
        Training Batch Size & $64$ \\
        Eval Batch Size & $8$ \\
        Learning Rate & $0.00001$ \\
        Weight Decay & $0.01$ \\
        \# Training Epochs & $7$ \\
        Max Sentence Length & $128$ \\
        Max gradient norm & $1$\\
        \bottomrule
    \end{tabular}
    \caption{Hyperparameter details for WSD model.}
    \label{tab:hyper-wsd}
\end{table}

\subsection{TA{\small BS}}
\label{implementation-details-tabs}

TA{\small BS} \cite{li-etal-2022-open} is an event type induction model, wherein the goal is to discover new event types without a pre-defined event ontology.
It utilizes two complementary trigger embedding spaces (mask view and token view) for classification.
To adapt this for ED, we follow the end-to-end event discovery setting in \cite{choi-etal-2022-tabs} while making the following modifications:
(1) \textbf{Dataset Composition}:
We utilize ACE \cite{doddington-etal-2004-automatic} dataset for training and development and our SPEED dataset for testing.
Our training data comprises 26 known event types from ACE, the validation set comprises 7 ACE event types, while our test set comprises 7 event types from \dataName{}.
(2) \textbf{Candidate Trigger Extraction}:
To improve trigger coverage, we extract all nouns and non-auxiliary verbs as candidate trigger mentions. 
(3) \textbf{Evaluation Setup}:
Trigger identification (\textbf{Tri-I}) F1 score is evaluated using the extracted candidate triggers.
For trigger classification (\textbf{Tri-C}), we first find the best cluster assignment of the predicted event clusters to the gold event types and then evaluate the F1 score.

We run our experiments for TA{\small BS} on an NVIDIA RTX 2080 Ti machine with support for 8 GPUs. The major hyperparameters for this model are listed in Table~\ref{tab:hyper-tabs}.

\begin{table}[h]
    \centering
    \small
    \begin{tabular}{lr}
        \toprule
        Pre-trained LM & BERT-Base \\
        Training Batch Size & $8$ \\
        Eval Batch Size & $8$ \\
        Gradient Accumulation Steps & $2$ \\
        Learning Rate & $0.00005$ \\
        Gradient Clipping & $1$ \\
        \# Pretrain Epochs & $10$ \\
        \# Training Epochs & $30$ \\
        Consistency Loss Weight & $0.2$ \\
        \# Target Unknown Event Types & 30 \\
        \bottomrule
    \end{tabular}
    \caption{Hyperparameter details for TA{\small BS} model.}
    \label{tab:hyper-tabs}
\end{table}

\subsection{ETypeClus}
\label{implementation-details-etypeclus}

ETypeClus \cite{shen-etal-2021-corpus} extracts salient predicate-object pairs and clusters their embeddings in a spherical latent space.
For consistency across our evaluations, we follow the re-implementation of the ETypeClus model in \cite{choi-etal-2022-tabs}, which consists of the latent space clustering stage of the ETypeClus pipeline and uses the embeddings of trigger mentions to be the input features.
We utilize the contextualized embeddings of the candidate triggers extracted from \dataName{} for unsupervised training.
The candidate trigger extraction process and the evaluation setup are the same as described in \S~\ref{implementation-details-tabs}.

We run our experiments for ETypeClus on an NVIDIA RTX 2080 Ti machine with support for 8 GPUs. The major hyperparameters for this model are listed in Table~\ref{tab:hyper-etypeclus}.

\begin{table}[h]
    \centering
    \small
    \begin{tabular}{lr}
        \toprule
        Pre-trained LM & BERT-Base \\
        Training Batch Size & $64$ \\
        Eval Batch Size & $64$ \\
        Learning Rate & $0.0001$ \\
        Gradient Clipping & $1$ \\
        \# Pretrain Epochs & $10$ \\
        \# Training Epochs & $50$ \\
        KL Loss Weight & $5$ \\
        Temperature & $0.1$\\
        \# Target Unknown Event Types & 30 \\
        \bottomrule
    \end{tabular}
    \caption{Hyperparameter details for ETypeClus model.}
    \label{tab:hyper-etypeclus}
\end{table}

\subsection{BERT-QA}

BERT-QA \cite{du-cardie-2020-event} is a classification model utilizing label semantics by formulating event detection as a question-answering task.
We run our experiments for BERT-QA on an NVIDIA RTX A6000 machine with support for 8 GPUs. The major hyperparameters for this model are listed in Table~\ref{tab:hyper-bert-qa}.

\begin{table}[h]
    \centering
    \small
    \begin{tabular}{lr}
        \toprule
        Pre-trained LM & RoBERTa-Large \\
        Training Batch Size & $6$ \\
        Eval Batch Size & $12$ \\
        Learning Rate & $0.001$ \\
        Weight Decay & $0.001$ \\
        Gradient Clipping & $5$ \\
        Training Epochs & $30$ \\
        Warmup Epochs & $5$ \\
        Max Sequence Length & $175$ \\
        Linear Layer Dropout & $0.2$ \\
        \bottomrule
    \end{tabular}
    \caption{Hyperparameter details for BERT\_QA model.}
    \label{tab:hyper-bert-qa}
\end{table}

\subsection{DEGREE}

DEGREE \cite{hsu-etal-2022-degree} is a generation-based model prompting using natural language templates.
We run our experiments for DEGREE on an NVIDIA RTX A6000 machine with support for 8 GPUs. The major hyperparameters for this model are listed in Table~\ref{tab:hyper-degree}.

\begin{table}[h]
    \centering
    \small
    \begin{tabular}{lr}
        \toprule
        Pre-trained LM & BART-Large \\
        Training Batch Size & $32$ \\
        Eval Batch Size & $32$ \\
        Learning Rate & $0.00001$ \\
        Weight Decay & $0.00001$ \\
        Gradient Clipping & $5$ \\
        Training Epochs & $45$ \\
        Warmup Epochs & $5$ \\
        Max Sequence Length & $250$ \\
        Max Output Length & $20$ \\
        Negative Samples & $15$ \\
        Beam Size & $1$ \\
        \bottomrule
    \end{tabular}
    \caption{Hyperparameter details for DEGREE model.}
    \label{tab:hyper-degree}
\end{table}

\subsection{TagPrime}

TagPrime \cite{hsu-etal-2023-simple} is a sequence tagger priming words to input text to convey more task-specific information.
We run our experiments for TagPrime on an NVIDIA RTX A6000 machine with support for 8 GPUs. The major hyperparameters for this model are listed in Table~\ref{tab:hyper-tagprime}.

\begin{table}[h]
    \centering
    \small
    \begin{tabular}{lr}
        \toprule
        Pre-trained LM & RoBERTa-Large \\
        Training Batch Size & $64$ \\
        Eval Batch Size & $8$ \\
        Learning Rate & $0.001$ \\
        Weight Decay & $0.001$ \\
        Gradient Clipping & $5$ \\
        Training Epochs & $100$ \\
        Warmup Epochs & $5$ \\
        Max Sequence Length & $175$ \\
        Linear Layer Dropout & $0.2$ \\
        \bottomrule
    \end{tabular}
    \caption{Hyperparameter details for TagPrime model.}
    \label{tab:hyper-tagprime}
\end{table}

\subsection{DyGIE++}

DyGIE++ \cite{wadden-etal-2019-entity} is a multi-task classification-based model utilizing local and global context via span graph propagation.
We run our experiments for DyGIE++ on an NVIDIA RTX A6000 machine with support for 8 GPUs. The major hyperparameters for this model are listed in Table~\ref{tab:hyper-dygie}.

\begin{table}[h]
    \centering
    \small
    \begin{tabular}{lr}
        \toprule
        Pre-trained LM & RoBERTa-Large \\
        Training Batch Size & $6$ \\
        Eval Batch Size & $12$ \\
        Learning Rate & $0.001$ \\
        Weight Decay & $0.001$ \\
        Gradient Clipping & $5$ \\
        Training Epochs & $60$ \\
        Warmup Epochs & $5$ \\
        Max Sequence Length & $200$ \\
        Linear Layer Dropout & $0.4$ \\
        \bottomrule
    \end{tabular}
    \caption{Hyperparameter details for DyGIE++ model.}
    \label{tab:hyper-dygie}
\end{table}

\subsection{Keyword}

This baseline model basically curates a list of keywords specific to each event and predicts a trigger for a particular event if it matches one of the curated event keywords.
Event keywords are curated by expert annotators based on the gold triggers appearing in the \dataName{} dataset and classified as high confidence, medium confidence, and low confidence based on their occurrence counts and false positive rates (as described in \S~\ref{sec:initial-event-analysis}.\footnote{We will release the set of keywords with our final code.}
Although this baseline accesses gold test data, it is meant to be a baseline to provide the upper cap for models of this family.

\subsection{GPT-3}

\begin{figure}[t]
    \centering
    \includegraphics[width=\columnwidth]{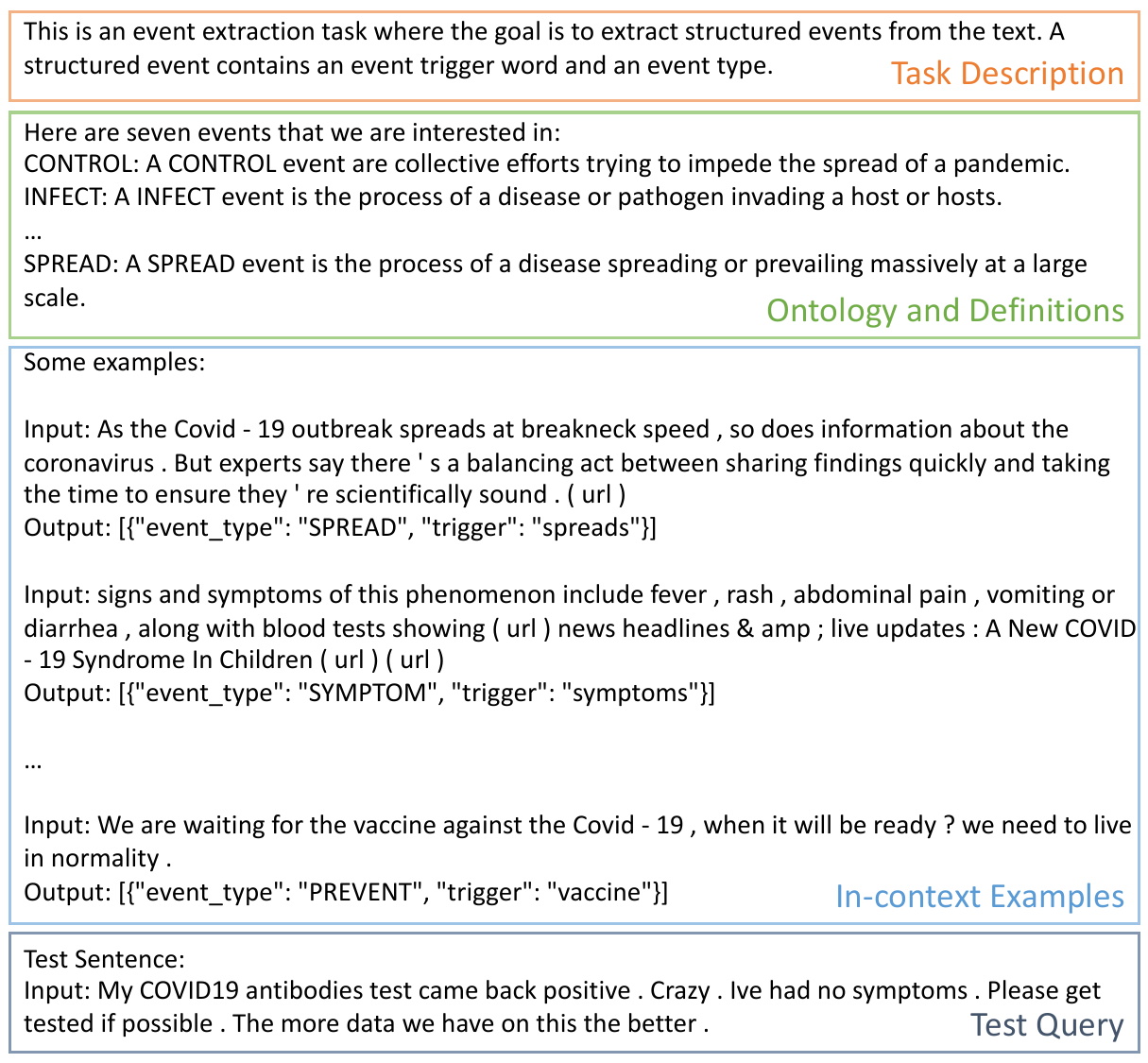}
    \caption{Illustration of the prompt used for GPT-3 model. It includes a task description, followed by ontology details of event types and their definitions. Next, we show some in-context examples for each event type and finally, provide the test sentence.}
    \label{fig:ed-gpt-prompt}
\end{figure}

We use the GPT-3.5 turbo model as the base GPT model.
We experiment with ChatGPT \cite{chatgpt} for tuning our prompts that ensure output consistency.
Our final prompt (as shown in Figure~\ref{fig:ed-gpt-prompt}) comprises a task definition, ontology details, 1 example for each event type, and the final test query.
We conducted a looser evaluation for GPT and only match if the predicted trigger text matches the gold trigger text (we didn't check the exact span match basically).

\section{Predicting Early Warnings for Monkeypox}
\label{sec:mpox-trends-appendix}




\subsection{Event-wise Analysis}

\begin{figure}[t]
    \centering
    \includegraphics[width=\columnwidth]{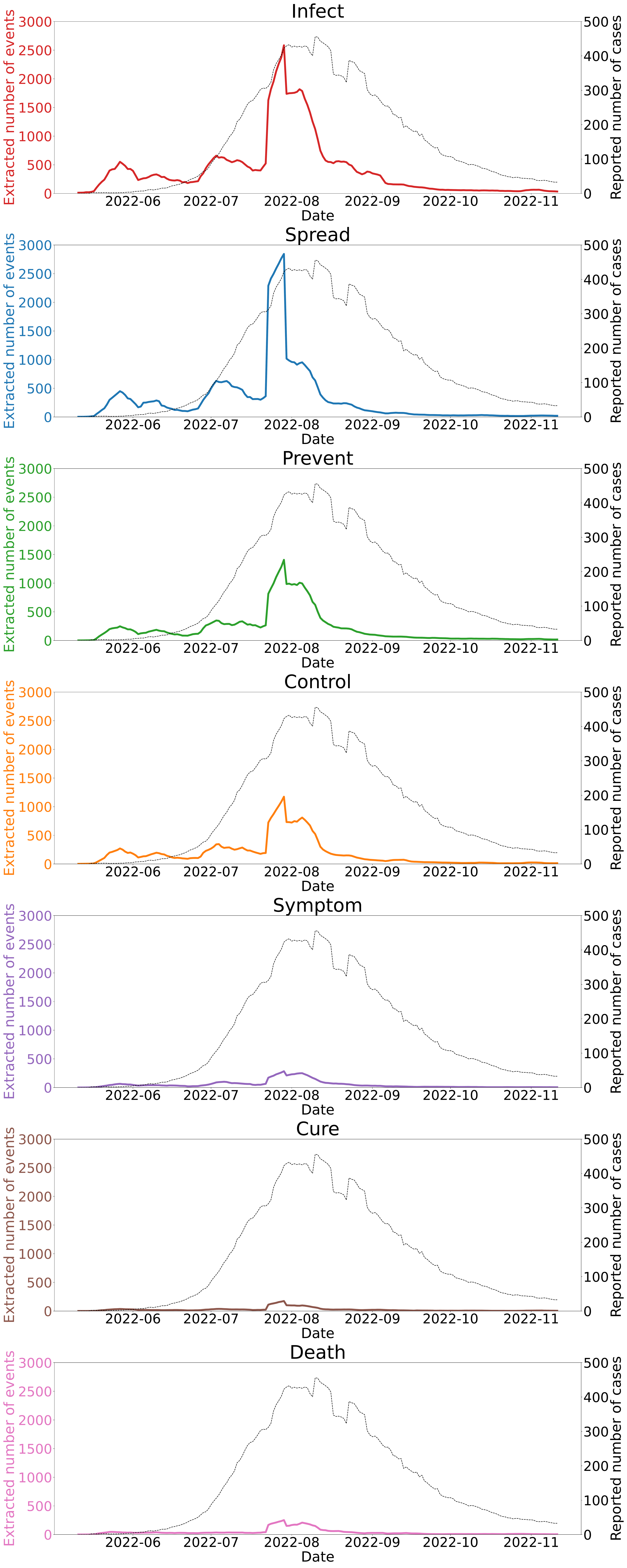}
    \caption{Number of reported Monkeypox cases and the number of extracted events for each SPEED event type from our trained BERT-QA model from $XX$ to $XX$}
    \label{fig:mpox-eventwise-analysis}
\end{figure}

As BERT-QA yields the strongest early warning signal (shown in Figure~\ref{fig:mpox-4-ED-models}), we conduct an analysis at a more granular level on the contribution of each event type to the early warning signal based on the trained BERT-QA output. 
We present the results in Figure \ref{fig:mpox-eventwise-analysis}, which leads to the following observations: 
(1) \textbf{Strength of indication varies among event types}: As indicated in Figure \ref{fig:mpox-eventwise-analysis}, event type \textit{infect} and \textit{spread} are strong indicators of the incoming surge in reported cases, while event type \textit{prevent} and \textit{control} can serve as indicators of medium strength. Event type \textit{symptom}, \textit{cure}, and \textit{death} are weak indicators that barely contribute to the early warning signal.
(2) \textbf{Distribution across event types can potentially reveal high-level disease characteristics}: We can infer some properties of diseases based on the frequency of mentions about particular events. For example, \textit{death} is less mentioned, which can indicate that \textit{Monkeypox} is less fatal compared to other epidemics like COVID.
We would like to mention that these are hypothetical properties based on predictions of our best model (which can be imperfect) and should be taken with a pinch of salt.

\end{document}